# Structured Prompt Language: Declarative Context Management for LLMs

Wen G. Gong

`wen.gong.research@gmail.com`

February 22, 2026

**Abstract**

We present **SPL** (Structured Prompt Language), a declarative SQL-inspired language that treats large language models as *generative knowledge bases* and their context windows as constrained resources. SPL provides explicit `WITH BUDGET/LIMIT` token management, an automatic query optimizer, `EXPLAIN` transparency analogous to SQL's `EXPLAIN ANALYZE`, and native integration of retrieval-augmented generation (RAG) and persistent memory in a single declarative framework. **SPL-flow** extends SPL into resilient agentic pipelines with a three-tier provider fallback strategy (Ollama → OpenRouter → self-healing retry) fully transparent to the `.spl` script.

Five extensions demonstrate the paradigm's breadth: (1) *Text2SPL* (multilingual NL→SPL translation); (2) *Mixture-of-Models* (MoM) routing that dispatches each `PROMPT` to a domain-specialist model at runtime; (3) *Logical Chunking*, an intelligent strategy for documents exceeding a single context window—expressed naturally through SPL's existing CTE syntax with no new constructs, decomposing a large query into a Map-Reduce pipeline that reduces attention cost from $O(N^2)$ to $O(N^2/k)$ and runs identically on cloud (parallel) or local hardware (sequential); (4) *SPL-flow*, a declarative agentic orchestration layer with resilient three-tier provider fallback; and (5) *BENCHMARK* for parallel multi-model comparison with automatic winner persistence.

We provide a formal EBNF grammar, two pip-installable Python packages (`spl-llm`, `spl-flow`), and comparison against Prompty, DSPy, and LMQL. SPL reduces prompt boilerplate by 65% on average, surfaces a 68× cost spread across model tiers as a pre-execution signal, and runs the identical `.spl` script at $0.002 on OpenRouter or at zero marginal cost on a local Ollama instance—without modification.

## 1 Introduction

The emergence of large language models (LLMs) [Brown et al., 2020, Vaswani et al., 2017] as general-purpose reasoning engines has created an unprecedented challenge in software engineering: *how to systematically manage the context window*—the fixed-size input buffer that determines what information an LLM can consider when generating responses. Context window sizes range from 4K tokens in early models to over 1M tokens in recent architectures, yet the challenge remains fundamentally the same: context is a *constrained resource* that must be allocated efficiently across competing demands.

The current state of LLM prompt engineering bears a striking resemblance to database programming before SQL. In the pre-SQL era (1960s), programmers wrote imperative, procedural code to navigate data structures—each database vendor had its own access methods, and there was no standard way to express *what* data was needed versus *how* to retrieve it. Codd's relational model [Codd, 1970] and the subsequent development of SQL [Chamberlin and Boyce, 1974] resolved this chaos by introducing a declarative language that separated logical queries from physical execution.

Today's prompt engineers face an analogous problem:

- <u>Manual token counting</u>: Developers hand-calculate whether their prompts fit within context windows, often through trial and error.
- <u>Ad hoc truncation</u>: When context exceeds limits, engineers apply crude truncation strategies with no systematic optimization.
- <u>No visibility</u>: There is no standard way to inspect how tokens are allocated across different parts of a prompt.
- <u>Provider lock-in</u>: Prompt code is typically tied to specific LLM APIs, making migration costly.
- <u>No composability</u>: Unlike SQL's Common Table Expressions (CTEs) and views, prompts are monolithic and resist modular reuse.

We propose **SPL (Structured Prompt Language)**[1], a declarative query language that applies the principles that made SQL successful—declarative specification, automatic optimization, composability, portability, and transparency—to the domain of LLM context management.

**A key distinction from prompt-content optimization frameworks such as DSPy:** SPL manages the token *budget*—the container that constrains how much context an LLM receives—not the prompt *content* itself. What information to include is the developer's concern; how much of it fits within the budget, and how to compress it when it does not, is SPL's responsibility.

SPL's central conceptual innovation is treating an LLM not merely as a function that maps text to text, but as a *generative knowledge base*—a system that not only stores but actively synthesizes knowledge. Table 1 previews the SQL–SPL parallel; Section 2 develops the full framework and its implications.

Table 1: The SQL–SPL parallel: from static to generative knowledge bases.

| Dimension | SQL (1970) | SPL (2026) |
| --- | --- | --- |
| Knowledge base | Static (rows/tables) | Generative (trained weights) |
| Query result | Retrieved data | Synthesized content |
| Constrained resource | Disk I/O, RAM | Token budget (context window) |
| Optimization goal | Minimize I/O | Minimize tokens, maximize relevance |
| `SELECT` | Retrieves existing data | Gathers context from sources |
| `GENERATE` | N/A | Creates new content from context |
| `EXPLAIN` | Shows query execution plan | Shows token allocation plan |
| CTEs / Views | Compose complex queries | Compose complex prompts |
| Transactions | ACID guarantees | Memory persistence + caching |
| Provider portability | Oracle, Postgres, MySQL | OpenRouter, Claude, GPT, Llama |

## 1.1 Contributions

This paper makes the following contributions:

- <u>Language design</u>: We present the complete syntax and formal grammar (EBNF) of SPL, a declarative language for LLM context management with explicit token budgeting, composable CTEs, user-defined functions, and integrated retrieval-augmented generation (RAG [Lewis et al., 2020]) and memory.

---

[1] An unrelated project sharing the acronym "SPL" uses XML-based template markup for agent instructions [Eshraw, 2025]; it is a template language, not a query language. This work's VS Code extension is at `https://github.com/digital-duck/SPL-LLM`.

- Context-as-resource model: We formalize the LLM context window as a constrained resource and present a token budget allocation algorithm analogous to query optimization in relational databases.
- EXPLAIN mechanism: We introduce transparent execution plans for LLM queries, enabling developers to inspect and reason about token allocation before execution.
- Reference implementation: We provide a complete, pip-installable Python package (`spl-llm`) with a hand-written recursive descent parser, optimizer, and provider-agnostic execution engine.
- Comparative analysis: We systematically compare SPL against existing prompt management approaches (Prompty, DSPy, LMQL) across key dimensions.
- SPL ecosystem: We describe five extensions that demonstrate SPL as an application platform, not merely a research prototype: (1) Text2SPL (multilingual NL interface), (2) Mixture-of-Models (MoM) routing, (3) Logical Chunking (declarative Map-Reduce for long contexts, powered by SPL's native CTE syntax), (4) SPL-flow (resilient agentic orchestration), and (5) `BENCHMARK` with automatic winner selection.

# 2   LLM as Generative Knowledge Base

The conceptual foundation of SPL is treating LLMs as *generative knowledge bases*. This perspective is not purely theoretical: practitioners who build AI applications daily find that the mental model of "LLM as database"—a system you query, whose responses depend on what you place in its context window, and whose implicit "tables" are its trained weights—is the most productive frame for understanding what LLMs actually do. SPL gives this practitioner intuition a formal, executable expression.

## 2.1   From Retrieval to Synthesis

Traditional information systems follow a retrieve-and-present paradigm: a user poses a query, the system searches its index, and returns matching documents. The user must then synthesize understanding from the retrieved items.

LLMs break this paradigm. When queried, an LLM does not merely retrieve stored information; it *synthesizes* a novel response by combining:
- Parametric knowledge: Information encoded in model weights during training
- Contextual knowledge: Information provided in the current prompt (the context window)
- Reasoning: Emergent multi-step inference capabilities

SPL's dual-clause architecture (`SELECT` + `GENERATE`) directly mirrors this distinction:
- `SELECT` populates the contextual knowledge (item 2 above), optionally structured by reasoning scaffolds (item 3)
- `GENERATE` triggers the full synthesis process, drawing on all three sources: parametric knowledge already encoded in the model weights (item 1), the contextual knowledge assembled by `SELECT` (item 2), and emergent reasoning (item 3)

This is why SPL is not merely "a prompt template language"—it is a *query language for a generative knowledge base*, where the "database" (LLM weights) can produce outputs that never existed in its "tables" (training data).

## 2.2   The Duality with SQL

In SQL, the `SELECT` statement both specifies what to retrieve *and* produces the result. In SPL, we separate these concerns:
- `SELECT` specifies what context to *gather* (retrieval phase)
- `GENERATE` specifies what to *create* from that context (synthesis phase)

This separation enables the optimizer to treat context gathering and generation as distinct phases with different resource characteristics, leading to better budget allocation.

*Note for SQL practitioners*: in SQL, `SELECT` is the statement that both gathers *and* produces the final output. In SPL, this output role belongs to `GENERATE`—the clause that triggers synthesis and delivers the response. SPL's `SELECT` is analogous to a SQL sub-query or `WITH` clause: it assembles context for the generation step but produces no output of its own.

## 2.3 Implications for Software Engineering

Treating LLMs as generative knowledge bases has several implications for software engineering practice:

- Prompt engineering becomes query engineering: The skill set shifts from "crafting text that tricks the model" to "declaring what context and what output are needed."
- Observability becomes possible: Just as SQL `EXPLAIN` enables database performance tuning, SPL `EXPLAIN` enables prompt performance tuning.
- Testing becomes structured: SPL queries can be validated statically (parse, analyze) and their resource usage estimated before any LLM call.
- Version control becomes meaningful: `.spl` files are human-readable, diffable, and reviewable—unlike embedded prompt strings in application code.

# 3 Background and Motivation

## 3.1 The Prompt Engineering Problem

As LLMs have become central to software systems, prompt engineering has emerged as a critical discipline. Developers must construct input sequences that effectively utilize the model's capabilities while respecting hard constraints on context window size and soft constraints on cost and latency.

Current prompt engineering practices are overwhelmingly *imperative*: developers write Python (or JavaScript, etc.) code that concatenates strings, counts tokens, applies heuristic truncation, and makes API calls [Wei et al., 2022]. This imperative approach has several well-documented problems:

- Tight coupling: Prompt construction logic is interleaved with application logic, making prompts difficult to test, version, and share.
- No budget discipline: Without explicit budget allocation, developers discover token overflows only at runtime (often in production).
- Optimization blind spots: Imperative code obscures the structure of token allocation, making it impossible to reason about whether the budget is spent optimally.
- Reinvention: Common patterns (RAG context injection, conversation history management, response caching) are reimplemented from scratch in every project.

These patterns were directly encountered during the development of Data-Copilot [Gong, 2024], a RAG-powered conversational analytics application, and motivated the design of SPL.

## 3.2 Lessons from SQL's Success

SQL has dominated database interaction for over five decades because it embodies a set of principles that proved universally valuable:

- Declarative: Specify *what* you want, not *how* to get it. The optimizer handles execution strategy.
- Optimizable: A query planner can reorder joins, choose indexes, and parallelize—impossible with imperative navigation.

- Composable: CTEs, views, subqueries, and functions enable modular construction of complex queries.
- Portable: The same SQL runs (with minor dialect differences) on Oracle, PostgreSQL, MySQL, SQLite, and hundreds of other engines.
- Transparent: `EXPLAIN` reveals the execution plan, enabling informed optimization by the developer.

SPL applies each of these principles to LLM context management, as we detail in the following sections.

## 3.3 The Token Budget as Constrained Resource

We propose a formal model of the LLM context window as a constrained resource allocation problem. Given a context window of $B$ tokens, an SPL query must allocate this budget across $n$ context sources $s_1, \ldots, s_n$, a generation task $g$, and a safety buffer $\beta$:

$$\sum_{i=1}^{n} \text{tokens}(s_i) + \text{tokens}(g) + \beta \leq B \tag{1}$$

Each source $s_i$ has:
- An *estimated size* $\hat{t}_i$ (tokens before optimization)
- An optional *limit* $L_i$ (the `LIMIT` clause)
- A *priority* $p_i$ (derived from source type: memory > cache > RAG > context)
- A *compression ratio* $c_i \in (0, 1]$ (applied when over budget)

The optimizer's task is to find allocations $a_i \leq \min(\hat{t}_i, L_i)$ that satisfy Equation 1 while maximizing information utility. This is analogous to the query optimizer in relational databases, which minimizes I/O cost subject to producing correct results.

# 4 The SPL Language

## 4.1 Design Philosophy

SPL follows three design principles:
- SQL familiarity: Leverage the muscle memory of millions of SQL developers. Anyone who can write `SELECT ... FROM ... WHERE` can learn SPL quickly.
- Budget-first: Every SPL query explicitly states its token budget. This forces developers to think about context allocation upfront, preventing runtime surprises.
- Source-agnostic: Context sources (user input, RAG results, memory) are accessed through a uniform interface, making queries composable regardless of where data originates.

## 4.2 Core Syntax

The fundamental SPL statement is the `PROMPT` query, which has the following structure:

```
PROMPT answer_question
WITH BUDGET 8000 tokens
USING MODEL claude-sonnet-4-5

SELECT
    system_role("You are a knowledgeable assistant"),
    context.question AS question LIMIT 200 tokens,
    rag.query("relevant docs", top_k=5) AS docs LIMIT 3000 tokens,
    memory.get("history") AS history LIMIT 500 tokens

WHERE
    docs.relevance > 0.7
```

```
ORDER BY
    docs.relevance DESC

GENERATE
    detailed_answer(question, docs, history)
WITH OUTPUT BUDGET 2000 tokens, TEMPERATURE 0.3, FORMAT markdown;
```

Listing 1: Basic SPL query structure

The key clauses are:

**PROMPT `<name>`** Names the query for reuse and reference.
**WITH BUDGET `<n>` tokens** Declares the total token budget for the entire operation (input + output).
**USING MODEL `<model>`** Specifies the target LLM, enabling model-specific token counting and cost estimation.
**SELECT** Gathers context from heterogeneous sources, each with an optional `LIMIT` cap.
**WHERE** Filters context items based on conditions (e.g., relevance thresholds for RAG results).
**ORDER BY** Orders context items, controlling which appear first in the assembled prompt.
**GENERATE** Specifies the generation task with output parameters.
**STORE RESULT IN `memory.<key>`** Persists results to the memory store for future queries.

### 4.3 Built-in Context Sources

SPL provides four built-in context source types:

Table 2: SPL built-in context sources.

| Source | Syntax | Description |
| --- | --- | --- |
| System role | `system_role("...")` | Sets the system/instruction prompt |
| Context | `context.<field>` | References external input parameters |
| RAG | `rag.query("...", top_k=N)` | Semantic retrieval from vector store |
| Memory | `memory.get("key")` | Key-value retrieval from persistent store |

### 4.4 Common Table Expressions (CTEs)

Like SQL, SPL supports CTEs via the `WITH` clause, enabling composition of complex prompts from simpler building blocks:

```
PROMPT analyze_and_recommend
WITH BUDGET 12000 tokens
USING MODEL claude-sonnet-4-5

WITH compressed_profile AS (
    SELECT
        context.user_profile AS profile
    LIMIT 500 tokens
),
relevant_docs AS (
    SELECT
        rag.query("product recommendations", top_k=3) AS docs
    LIMIT 2000 tokens
)
```

```
SELECT
    system_role("You are a product recommendation expert"),
    compressed_profile AS profile,
    relevant_docs AS docs,
    memory.get("purchase_history") AS history LIMIT 1000 tokens

GENERATE
    product_recommendations(profile, docs, history)
WITH OUTPUT BUDGET 4000 tokens, TEMPERATURE 0.5

STORE RESULT IN memory.last_recommendations;
```

Listing 2: CTEs for composable prompt construction

CTEs serve two purposes: (1) they enable token-limited preprocessing of context sources, and (2) they create reusable named components within a query.

## 4.5 User-Defined Functions

SPL supports `CREATE FUNCTION` for reusable prompt components:

```
CREATE FUNCTION compress_bio(bio, max_tokens)
RETURNS text
AS $$
    SELECT bio LIMIT max_tokens tokens
$$;
```

Listing 3: User-defined function

Functions encapsulate common context processing patterns (compression, formatting, extraction) and can be invoked from any `SELECT` or `GENERATE` clause.

## 4.6 EXPLAIN Mechanism

The `EXPLAIN` statement produces a detailed token allocation plan without executing the query:

```
EXPLAIN PROMPT answer_question;
```

Listing 4: EXPLAIN for execution plan transparency

This produces output such as:

```
Execution Plan for: answer_question
============================================================
Budget: 8,000 tokens | Model: claude-sonnet-4-5

Token Allocation:
+-- __system_role__           20 tokens  (  0.2%)
+-- history                  500 tokens  (  6.2%)  [from memory]
+-- docs                   3,000 tokens  ( 37.5%)  [cache MISS]
+-- question                 200 tokens  (  2.5%)
+-- Output Budget          2,000 tokens  ( 25.0%)
\-- Buffer                 2,280 tokens  ( 28.5%)
                          ----------
Total                     5,720 / 8,000 tokens (71.5%)

Estimated Cost: $0.0412
```

The `EXPLAIN` output provides: (1) a tree view of token allocation across all context sources, (2) the percentage of budget consumed by each source, (3) cache status for each source, (4) compression annotations where applicable, and (5) estimated cost based on the target model's pricing.

This mechanism is directly analogous to SQL's `EXPLAIN` / `EXPLAIN ANALYZE`, which shows query execution plans including join strategies, index usage, and estimated row counts. For LLM queries, the "execution plan" is the token allocation strategy.

### 4.7 Formal Grammar

We provide a complete EBNF grammar for SPL. The grammar is context-free and LL(1)-parseable by a recursive descent parser. The full grammar (all productions including `WHERE`, `ORDER BY`, `CREATE FUNCTION`, `EXECUTE`, expressions, and terminals) is reproduced in Appendix F and is also available as `docs/grammar.ebnf` in the project repository at https://github.com/digital-duck/SPL. The core `PROMPT` statement structure is:

Listing 5: SPL grammar — core productions (see Appendix F for full EBNF)
```
program     = statement (";" statement)* ";"? EOF ;
statement   = prompt_stmt | create_func_stmt | explain_stmt | execute_stmt
   ;

prompt_stmt = "PROMPT" IDENT
              ("WITH" "BUDGET" INTEGER "TOKENS")?
              ("USING" "MODEL" (IDENT | STRING))?
              ("CACHE" "FOR" INTEGER IDENT)?
              cte_block?
              "SELECT" select_item ("," select_item)*
              ("WHERE" condition (("AND" | "OR") condition)*)?
              ("ORDER" "BY" order_item ("," order_item)*)?
              "GENERATE" IDENT "(" arg_list? ")"
              ("WITH" generate_opt ("," generate_opt)*)?
              ("STORE" "RESULT" "IN" "memory" "." IDENT)? ;
```

## 5 Context as a Constrained Resource

### 5.1 Information-Theoretic Formulation

We model an SPL query as a resource allocation problem. Let $Q$ be an SPL query with total budget $B$, output budget $O$, and context sources $S = \{s_1, \ldots, s_n\}$. The *input budget* available for context is:

$$B_{\text{input}} = B - O - \beta \qquad (2)$$

where $\beta$ is a safety buffer (typically 5–10% of $B$).

Each context source $s_i$ has:
- Raw token count: $t_i = \text{tokens}(s_i)$
- User-specified limit: $L_i$ (from `LIMIT` clause, or $\infty$ if unspecified)
- Effective allocation: $a_i = \min(t_i, L_i)$

The budget constraint is:

$$\sum_{i=1}^{n} a_i \leq B_{\text{input}} \qquad (3)$$

When $\sum a_i > B_{\text{input}}$ (over-budget), the optimizer applies compression. As a closed-form approximation, uniform proportional scaling gives:

$$a'_i = a_i \cdot \frac{B_{\text{input}}}{\sum_{j=1}^{n} a_j} \quad \text{(proportional scaling, simplified)} \tag{4}$$

The implemented procedure (Algorithm 1) refines this with a largest-first greedy strategy: sources are sorted by allocation size descending, and each is compressed by up to 50% of its allocation until the excess is eliminated. This concentrates reduction on the largest sources first, preserving small, fixed-size items (e.g., system role, short questions) in full. Note that compression is applied purely by size; the priority ordering (memory > cache > RAG > context) governs *execution sequencing* only, not allocation protection. The design rationale is deliberate: tying compression to priority would allow high-priority sources to consume their full allocation at the expense of lower-priority ones, potentially leaving some source types with zero tokens. Size-driven compression instead ensures that *all* sources contribute to the context proportionally—a large memory entry is compressed more than a short system role, but both remain present. Priority-based compression protection is a natural extension for future work (Section 10).

## 5.2 Cost Model

SPL's cost model parallels relational algebra's cost model. In SQL, the optimizer estimates I/O cost (disk pages read, index lookups). In SPL, the optimizer estimates:

$$\text{Cost}(\mathcal{Q}) = \text{cost\_per\_input\_token}(m) \cdot \sum a_i + \text{cost\_per\_output\_token}(m) \cdot O \tag{5}$$

where $m$ is the target model. This enables `EXPLAIN` to display estimated dollar cost before execution, allowing developers to make informed decisions about model selection and budget allocation.

## 5.3 Execution Order Optimization

The SPL optimizer also determines the order in which context sources are resolved, following a priority hierarchy:
- Cache hits: Previously computed results (zero latency, zero cost)
- Memory: SQLite key-value lookups (sub-millisecond, zero cost)
- RAG: FAISS/ChromaDB vector search (milliseconds, minimal cost)
- Context: External input parameters (variable latency)
- LLM calls: The generation step (highest latency and cost)

This ordering minimizes latency by executing the cheapest operations first, analogous to how SQL optimizers prefer index scans over full table scans. Priority affects *execution sequencing only*—it does not protect sources from compression, which is governed solely by allocation size (see Section 5).

# 6 The SPL Engine

## 6.1 Architecture

The SPL engine follows a classic compiler pipeline architecture, mirroring the structure of SQL query engines. The pipeline consists of five stages: **Lexer**, **Parser**, **Semantic Analyzer**, **Optimizer**, and **Executor**.

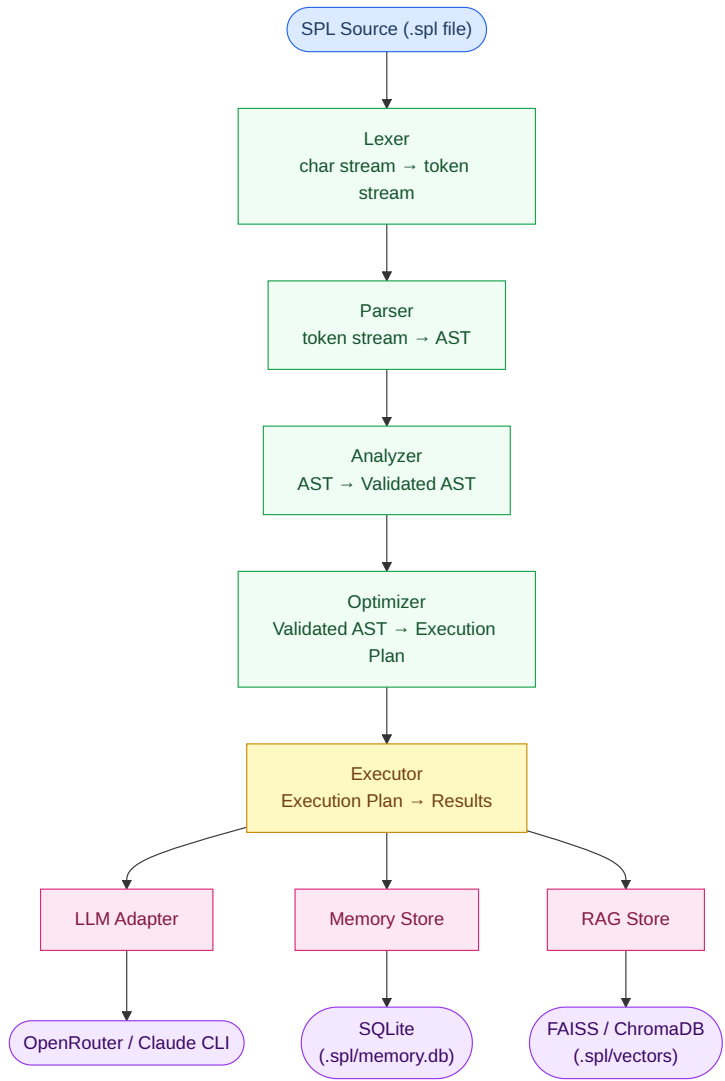

Figure 1: SPL engine pipeline architecture. The compiler stages (blue input, green pipeline) transform SPL source into an execution plan consumed by the executor (yellow), which dispatches to three back-end subsystems (pink): LLM adapters (OpenRouter / Claude CLI), a SQLite memory store, and a vector RAG store supporting both FAISS and ChromaDB (purple).

### 6.1.1 Lexer

The lexer performs character-by-character scanning, producing a stream of typed tokens. It handles: SQL-style keywords (case-insensitive), string literals (single and double quotes), numeric literals (integer and floating-point), dot notation for qualified references (`context.field`), `$$` delimiters for function bodies, and `--` single-line comments. The keyword set includes 30+ reserved words (Table 3).

Table 3: SPL keyword categories.

| Category | Keywords |
|---|---|
| Structural | `PROMPT, SELECT, GENERATE, WITH, AS, FROM` |
| Budget | `BUDGET, TOKENS, LIMIT, OUTPUT` |
| Model | `USING, MODEL, TEMPERATURE, FORMAT` |
| Filtering | `WHERE, AND, OR, NOT, IN, ORDER, BY, ASC, DESC` |
| Functions | `CREATE, FUNCTION, RETURNS` |
| Control | `EXPLAIN, EXECUTE, PARAMS, STORE, RESULT, CACHE, FOR` |
| Built-ins | `system_role, context, rag, memory` |

### 6.1.2 Parser

The parser is a hand-written recursive descent parser—a deliberate architectural choice over parser generators (ANTLR, Lark) for three reasons:

- <u>Zero dependencies</u>: The parser has no external dependencies, reducing the attack surface and installation friction.
- <u>Full control</u>: Error messages can reference SPL-specific concepts ("expected GENERATE clause after SELECT") rather than generic parse errors.
- <u>Formalizability</u>: The one-to-one correspondence between grammar productions and parser methods makes the implementation a direct proof of the grammar's LL(1) property, suitable for inclusion in a formal language specification.

The parser produces a typed AST with dataclass nodes for each syntactic construct (Table 4).

Table 4: Core AST node types.

| Node Type | Description |
|---|---|
| `PromptStatement` | Top-level query with budget, model, select, generate clauses |
| `SelectItem` | Single context source with optional alias and token limit |
| `CTEClause` | Common table expression (named subquery) |
| `GenerateClause` | Generation specification with parameters |
| `CreateFunctionStatement` | User-defined function definition |
| `ExplainStatement` | EXPLAIN query |
| `ExecuteStatement` | Parameterized query execution |

### 6.1.3 Semantic Analyzer

The analyzer performs validation passes on the AST:

- <u>Name resolution</u>: Verifies that all referenced aliases (in `GENERATE` arguments, `WHERE` conditions) correspond to defined `SELECT` items or CTEs.

- Budget arithmetic: Warns when the sum of `LIMIT` clauses exceeds the total budget.
- Type checking: Validates that context sources are used with correct syntax (e.g., `rag.query` requires a string argument).
- Duplicate detection: Reports duplicate prompt names, aliases, or CTE definitions.

#### 6.1.4 Optimizer

The optimizer transforms a validated AST into an `ExecutionPlan`—an ordered sequence of steps with token allocations. The optimization algorithm (Algorithm 1) proceeds as follows:

---
**Algorithm 1** SPL Token Budget Optimizer
---
**Require:** $B$: total budget, $O$: output budget, $S$: set of SELECT items
**Ensure:** ExecutionPlan with token allocations
1: $B_{\text{input}} \leftarrow B - O - \beta$
2: **for** each $s_i \in S$ **do**
3: $\quad \hat{t}_i \leftarrow \text{estimate\_tokens}(s_i)$
4: $\quad a_i \leftarrow \min(\hat{t}_i, L_i)$ $\quad\quad\quad\quad\quad\quad\quad\quad\quad\quad\quad\quad\quad\quad\quad\quad\quad\triangleright$ Apply LIMIT cap
5: **end for**
6: **if** $\sum a_i > B_{\text{input}}$ **then**
7: $\quad$ Sort $S$ by $a_i$ descending $\quad\quad\quad\quad\quad\quad\quad\quad\quad\quad\quad\quad\quad\triangleright$ Compress largest first
8: $\quad$ excess $\leftarrow \sum a_i - B_{\text{input}}$
9: $\quad$ **for** each $s_i$ in sorted order **do**
10: $\quad\quad$ reduction $\leftarrow \min(a_i \times 0.5, \text{excess})$
11: $\quad\quad a_i \leftarrow a_i - \text{reduction}$
12: $\quad\quad$ excess $\leftarrow$ excess $-$ reduction
13: $\quad\quad$ **if** excess $\leq 0$ **then**
14: $\quad\quad\quad$ **break**
15: $\quad\quad$ **end if**
16: $\quad$ **end for**
17: **end if**
18: Sort $S$ by priority: memory $>$ cache $>$ RAG $>$ context
19: **return** ExecutionPlan($S$, allocations, cost estimate)
---

#### 6.1.5 Executor

The executor processes the ExecutionPlan step by step:
- Resolve each context source (load from memory, execute RAG queries, read parameters)
- Apply token limits via truncation
- Check prompt cache (hash-based, with TTL)
- Assemble the final prompt string with system role, context sections, and generation instruction
- Call the LLM adapter
- Optionally store results via `STORE RESULT IN memory.<key>`
- Return structured result with content, token usage, timing, and cost

### 6.2 LLM Adapter Interface

SPL achieves provider-agnosticism through an abstract adapter interface:

Listing 6: LLM adapter interface (Python)
```
class LLMAdapter(ABC):
    @abstractmethod
```

```
    async def generate(
        self, prompt: str, model: str,
        max_tokens: int = 4096,
        temperature: float = 0.7,
        system: str | None = None,
    ) -> GenerationResult: ...

    @abstractmethod
    def count_tokens(self, text: str, model: str) -> int: ...

    @abstractmethod
    def list_models(self) -> list[str]: ...
```

The reference implementation includes four adapters spanning local and cloud deployment modes:
- Ollama Adapter (Local): Wraps the Ollama REST API (`localhost:11434`) to run open-weight models (Llama, Mistral, Phi, Qwen, etc.) entirely on-premises. This adapter is zero-cost beyond hardware and requires no internet connection, making it suitable for privacy-sensitive and air-gapped deployments.
- OpenRouter Adapter (Cloud aggregator): Calls the OpenRouter.ai [OpenRouter, 2024] API, which provides access to 100+ models (Claude, GPT, Llama, Mistral, Gemini, etc.) through a single unified endpoint. This realizes SPL's vision of provider-agnosticism at the infrastructure level.
- Direct Cloud Adapters (Cloud native): Dedicated adapters for Anthropic Claude, OpenAI GPT, and Google Gemini APIs call each provider's native endpoint directly, enabling use of provider-specific features (extended thinking, structured outputs, grounding) without routing through an aggregator.
- Claude CLI Adapter (Development): Wraps the Claude Code CLI via subprocess, leveraging subscription-based billing for zero-marginal-cost development (zero additional cost to the developer under a Claude Pro subscription). This adapter demonstrates that SPL can interface with any LLM access method, not just HTTP APIs.

### 6.3 Storage Layer

SPL integrates two storage systems, both file-based for portability:
- SQLite Memory Store (`.spl/memory.db`): Provides persistent key-value storage for `memory.get()` and `STORE RESULT IN memory.<key>`. Also stores prompt cache entries with TTL support. SQLite was chosen for its zero-configuration deployment, battle-tested reliability, and inclusion in the Python standard library.
- Vector Store (`.spl/vectors.*`): Provides semantic similarity search for `rag.query()`. The engine supports two backends: **FAISS** [Johnson et al., 2019] for file-based, zero-configuration deployment, and **ChromaDB** [Chroma Core, 2023] for richer metadata filtering and a managed-server mode. Both expose an identical Python interface, making the choice transparent to SPL queries. Document metadata is stored in a companion SQLite database (`.spl/vectors_meta.db`).

Both stores reside in a `.spl/` directory whose location follows the deployment mode: when running from a cloned source repository the directory is created at `$REPO_ROOT/.spl/` (co-located with the code, analogous to `.git/`); when installed via `pip install spl-llm` it defaults to `$HOME/.spl/` (a single user-level store shared across projects). The directory is gitignore-able and self-contained, enabling reproducible deployments.

# 7 Related Work

We compare SPL against existing approaches for managing LLM interactions. Table 5 provides a feature-level comparison; we discuss each approach below.

Table 5: Feature comparison of LLM prompt management approaches.

| ID | Feature | SPL | LMQL | Prompty | DSPy | LangChain |
|---|---|---|---|---|---|---|
| F1 | SQL knowledge reusable | ✓ | Partial[d] | – | – | – |
| F2 | SQL-like CTEs & functions | ✓ | – | – | – | – |
| F3 | Formal grammar (EBNF) | ✓ | ✓ | – | – | – |
| F4 | Zero-dependency parser | ✓ | – | N/A | N/A | N/A |
| F5 | Declarative query language | ✓ | ✓ | – | – | – |
| F6 | Explicit token budgeting | ✓ | Partial[a] | – | – | – |
| F7 | EXPLAIN execution plan | ✓ | – | – | – | – |
| F8 | Automatic compression | ✓ | – | – | – | – |
| F9 | Built-in RAG | ✓ | – | – | – | Via plugins |
| F10 | Persistent memory | ✓ | – | – | – | Via plugins |
| F11 | Provider-agnostic | ✓ | ✓ | Partial[b] | ✓ | ✓ |
| F12 | Prompt optimization | –[c] | – | – | ✓ | – |
| *Extensions & Applications* | | | | | | |
| F13 | Multilingual NL→query interface | ✓ | – | – | – | – |
| F14 | Mixture-of-Models routing | ✓ | – | – | – | Via code |
| F15 | Declarative Map-Reduce chunking | ✓ | – | – | – | Via code |
| F16 | Resilient agentic pipelines | ✓ | – | – | Partial | Via LangGraph[f] |
| F17 | Parallel model benchmarking | ✓ | – | – | Partial[e] | – |

[a] LMQL supports per-variable length constraints but not global budget allocation.
[b] Prompty primarily targets Azure/OpenAI; community adapters exist for others.
[c] SPL optimizes *token allocation* (resource management); DSPy optimizes *prompt content* (few-shot examples). These are complementary, not competing, concerns.
[d] LMQL borrows SQL keywords (`FROM`, `WHERE`) for token-level generation constraints, but its mental model differs substantially from relational query thinking. SPL is intentionally modeled on SQL end-to-end, enabling practitioners with SQL experience to adopt GenAI workflows without learning a new programming paradigm.
[e] DSPy supports evaluation and optimization loops for prompt content; SPL `BENCHMARK` compares model outputs, token costs, and latency in parallel, automatically persisting the winning model identifier for subsequent executions.
[f] LangGraph requires explicit graph construction with nodes, edges, and state management defined in application code. SPL-flow is fully declarative: provider fallback (Ollama → OpenRouter → self-healing retry) is automatic and transparent to the `.spl` script. DSPy supports module chaining but provides no provider-level fault tolerance.

## 7.1 LMQL

LMQL [Beurer-Kellner et al., 2023] (Language Model Query Language) is an SQL-inspired query language for constraining LLM generation. It supports variable-level constraints (e.g., `where len(ANSWER) < 100`), logit manipulation, and multi-model dispatch.

**Key differences from SPL:**
- *Constraint-focused vs. resource-focused*: LMQL focuses on constraining the *output* of LLM generation (format, length, valid tokens). SPL focuses on managing the *input*—how con-text is gathered, compressed, and allocated within the token budget. These are complementary: LMQL constrains what comes out, SPL manages what goes in.
- *No global budgeting*: LMQL supports per-variable token limits but has no concept of a global token budget with cross-component allocation.
- *No context management*: LMQL does not provide built-in RAG, memory, or context compression.
- *Research focus*: LMQL originated as a research project and has seen limited production adoption.

## 7.2 Prompty

Prompty [Microsoft, 2024] is Microsoft's approach to standardizing prompt engineering through a YAML-based file format (`.prompty`). Prompty files declare model parameters, input/output schemas, and template text with variable interpolation. It integrates with Microsoft's Semantic Kernel and Azure AI Foundry ecosystem.

**Key differences from SPL:**
- *Template vs. query*: Prompty is a *template format*—it defines static prompt structures with variable slots. SPL is a *query language*—it expresses dynamic context gathering, optimization, and generation as a single declarative query.
- *No token budgeting*: Prompty relies on the underlying model's `max_tokens` parameter and provides no mechanism for allocating tokens across prompt components.
- *No execution planning*: Prompty has no equivalent of `EXPLAIN` for inspecting token allocation.
- *No integrated RAG/memory*: Context sources (RAG, memory) must be managed externally and injected as variables.
- *Ecosystem coupling*: Prompty is designed primarily for the Azure/Semantic Kernel ecosystem, while SPL is provider-agnostic by design.

Prompty and SPL operate at different levels of abstraction: Prompty standardizes the *format* of prompt files, while SPL standardizes the *semantics* of context management as a query language.

## 7.3 DSPy

DSPy [Khattab et al., 2024] is a framework for "programming—not prompting—foundation models." It introduces a Python-embedded paradigm where LLM interactions are defined as typed `Signature` classes with input/output fields, composed using modules (`ChainOfThought`, `ReAct`, etc.), and optimized through automated prompt tuning.

**Key differences from SPL:**
- *Python-embedded vs. standalone language*: DSPy is a Python library; SPL is a standalone query language with its own grammar and parser. This makes DSPy easier to adopt for Python developers but harder to analyze statically, version independently, and port to other ecosystems.
- *Prompt optimization vs. resource optimization*: DSPy's key innovation is *automatic prompt optimization*—finding the best few-shot examples, instructions, and module configurations through search. SPL's key innovation is *resource optimization*—managing the token budget across heterogeneous context sources. These are *complementary* concerns: a DSPy-optimized prompt still needs to fit within a token budget, and an SPL query could benefit from DSPy-style content optimization within its `SELECT` clauses.
- *No token visibility*: DSPy does not expose token allocation to the developer. There is no equivalent of `EXPLAIN` or `WITH BUDGET`.
- *No integrated storage*: DSPy does not provide built-in memory or RAG; these must be integrated externally.

We view DSPy and SPL as addressing orthogonal aspects of the same problem. A future integration could use DSPy for optimizing the *content* of each `SELECT` item while using SPL for managing the *budget* across items.

## 7.4 Imperative Frameworks

LangChain [Chase, 2022], LlamaIndex [Liu, 2022], and Semantic Kernel [Microsoft, 2023] provide imperative APIs for LLM application development. While powerful and flexible, they require developers to manually manage token counting, context assembly, and truncation. They provide building blocks (vector stores, memory modules, prompt templates) but leave the orchestration to imperative code. More recently, LangChain Expression Language (LCEL) and LangGraph have introduced graph-based and declarative composition primitives, narrowing—but not eliminating—this gap; neither provides explicit token budget management or pre-execution EXPLAIN. Multi-agent orchestration frameworks such as AutoGen [Wu et al., 2023] (Microsoft) and CrewAI [CrewAI, Inc., 2024] extend this landscape with role-based agent composition and multi-agent conversation patterns. SPL-flow's declarative graph (Section 9.4) targets a similar goal—composing LLM agents into resilient pipelines—but routes entire `PROMPT` queries with explicit token budgets as first-class nodes, preserving resource-visibility guarantees at the orchestration layer that free-form message-passing frameworks do not provide. SPL's value proposition relative to all these frameworks is analogous to SQL's value relative to procedural database access: a declarative layer that enables optimization and transparency.

## 7.5 Budget-Aware LLM Optimization

Recent work has begun to treat LLM token consumption as a first-class optimization concern. Zhao et al. [Zhao et al., 2025] propose *access paths* for LLM-powered ordering operations, designing a budget-aware optimizer for `ORDER BY` semantics over LLM-generated rankings. Their work shares SPL's core intuition—that token budgets should be managed by a query optimizer rather than by application code—but targets a specific operator (ordering) rather than providing a general-purpose query language. SelfBudgeter [Li et al., 2025] addresses token efficiency from the output side, training models to dynamically allocate reasoning token budgets at inference time. SPL is complementary: it manages the *input* context budget declaratively before any LLM call, while SelfBudgeter optimizes how many tokens the model spends *generating* its answer.

## 7.6 Output Structuring and Constrained Decoding

Several frameworks focus on enforcing *output* structure rather than managing *input* context. Guidance [Microsoft Research, 2023] (Microsoft) and Outlines [Willard and Louf, 2023] enforce structured generation by constraining the decoder's logit distribution at token-generation time using context-free grammars or regular expressions. BAML and TypeChat take a schema-first approach, guaranteeing JSON-typed outputs via prompt engineering and validation loops. SPL's `FORMAT json` clause in `GENERATE` currently relies on model prompting for output structure; integrating a constrained-decoding backend (e.g. Outlines) would provide formal output guarantees and represents a natural future extension. These frameworks operate at the output boundary, while SPL operates at the input boundary—the two concerns are orthogonal and complementary.

## 7.7 Provider-Native Caching

Major LLM providers now offer server-side prompt caching for static prefixes—Anthropic's prompt caching [Anthropic, 2024], OpenAI's similar facility, and Google Gemini's context caching. SPL's `CACHE FOR` clause is a *client-side* semantic layer: it caches the full LLM response, keyed on the rendered prompt, for a user-specified duration. These two mechanisms are complementary—provider-side caching reduces the cost of repeated static *prefixes* within a session, while SPL's client-side caching eliminates redundant API calls entirely for identical queries across sessions. Future SPL-flow versions could detect cacheable static prefixes and activate provider-side caching automatically, capturing savings at both layers simultaneously.

# 8 Evaluation

We evaluate SPL through five experiments: (1) developer experience comparison against imperative Python, (2) token budget optimization behavior, (3) cross-model cost estimation, (4) EXPLAIN output showcase, and (5) systematic feature verification. All experiments run without LLM API calls—they exercise the parser, analyzer, optimizer, and EXPLAIN pipeline, demonstrating that SPL provides value *before* any tokens are spent.

## 8.1 Experiment 1: Developer Experience

We implement five benchmark tasks in both SPL and imperative Python, measuring code complexity and developer burden. The Python baselines use manual `tiktoken` [OpenAI, 2023] counting, explicit truncation, and hand-written caching—representative of real-world prompt engineering practice.

Table 6: Developer experience: SPL vs. imperative Python across five benchmark tasks.

| Task | SPL LoC | Python LoC | Reduction | Token Ops (SPL/Py) | Budget Visible | Static Validate |
|---|---|---|---|---|---|---|
| Simple QA | 9 | 20 | 55.0% | 0 / 4 | Yes / No | Yes / No |
| RAG-augmented QA | 17 | 51 | 66.7% | 0 / 9 | Yes / No | Yes / No |
| Multi-step CTE | 24 | 63 | 61.9% | 0 / 7 | Yes / No | Yes / No |
| Function Reuse | 16 | 43 | 62.8% | 0 / 11 | Yes / No | Yes / No |
| Cached Repeat | 9 | 42 | 78.6% | 0 / 4 | Yes / No | Yes / No |
| **Average** | **15** | **44** | **65.0%** | **0 / 7** | | |

SPL achieves a **65% average reduction in lines of code** while eliminating all 35 manual token-counting operations across the five tasks. (A *token operation* is any line of code that explicitly invokes a token-counting API, performs token arithmetic, or applies manual truncation logic—e.g., calls to `tiktoken.encode()`, `len(tokens) > limit`, or substring truncation based on a token count.) Every SPL query supports pre-execution budget visibility (`EXPLAIN`) and static validation (`spl validate`)—capabilities entirely absent from the imperative approach.

## 8.2 Experiment 2: Token Budget Optimization

We analyze the optimizer's behavior under varying budget levels and increasing context pressure.

### 8.2.1 Allocation Under Varying Budgets

Running the same query structure with budgets from 2K to 32K tokens:

The optimizer allocates tokens proportionally: RAG and memory scale with budget while system role and question remain fixed. At 32K, the RAG limit (8K) and memory limit (2K) cap out, leaving substantial buffer—visible to the developer via `EXPLAIN`.

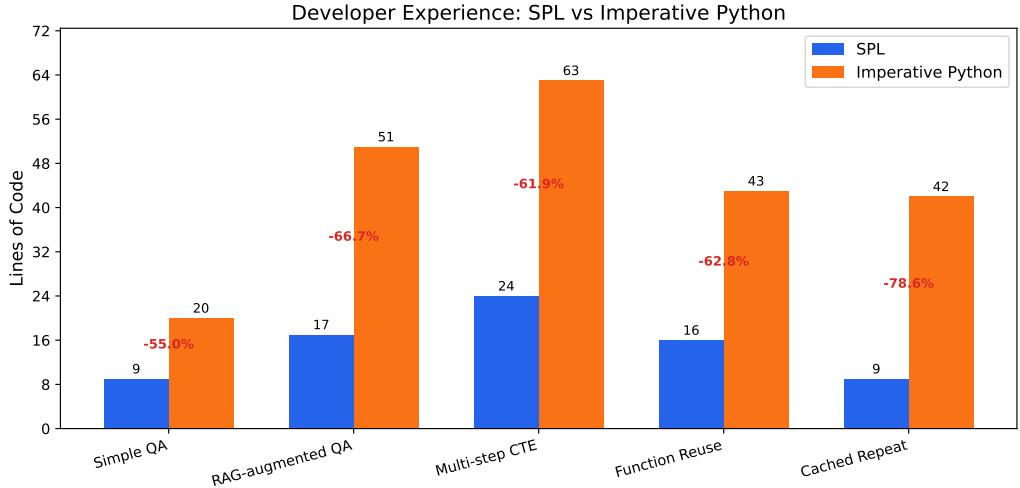

Figure 2: Lines of code comparison: SPL (blue) vs. imperative Python (orange). Percentage labels show code reduction. The most dramatic reduction (78.6%) occurs for cached queries, where SPL provides automatic caching while Python requires manual cache implementation.

Table 7: Token allocation strategy adapts proportionally to budget level.

| Budget | System | Question | RAG | Memory | Output | Buffer | Util% |
|---:|---:|---:|---:|---:|---:|---:|---:|
| 2,000 | 20 | 200 | 666 | 250 | 500 | 364 | 81.8% |
| 4,000 | 20 | 200 | 1,333 | 500 | 1,000 | 947 | 76.3% |
| 8,000 | 20 | 200 | 2,666 | 1,000 | 2,000 | 2,114 | 73.6% |
| 16,000 | 20 | 200 | 5,333 | 2,000 | 4,000 | 4,447 | 72.2% |
| 32,000 | 20 | 200 | 8,000 | 2,000 | 8,000 | 13,780 | 56.9% |

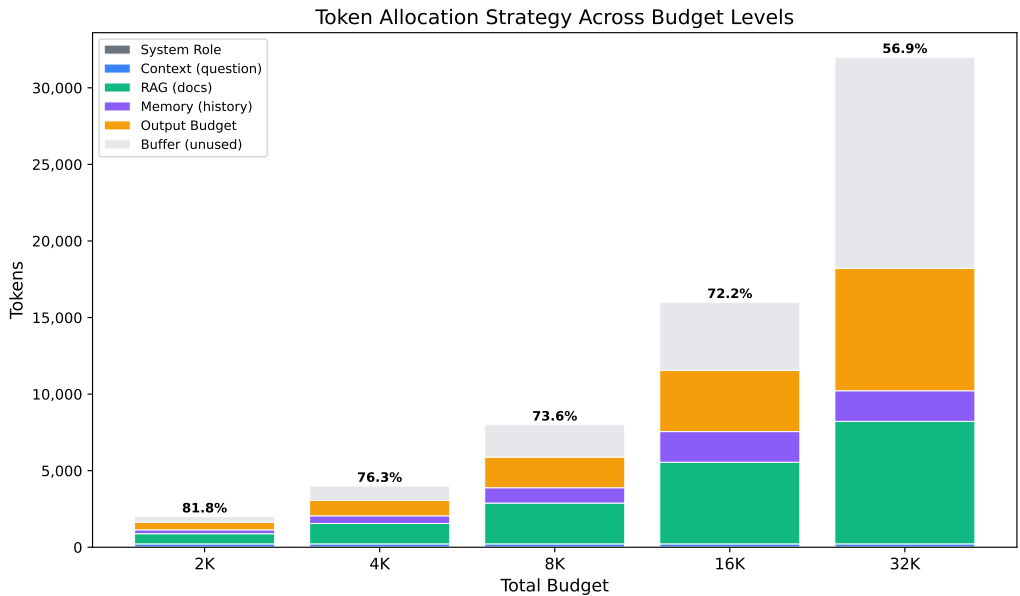

Figure 3: Stacked token allocation across budget levels. RAG (green) and memory (purple) scale proportionally; system role and question remain constant. Buffer (gray) grows at high budgets when source limits cap out.

### 8.2.2 Compression Under Budget Pressure

When context demands exceed the budget, the optimizer applies proportional compression (largest items first):

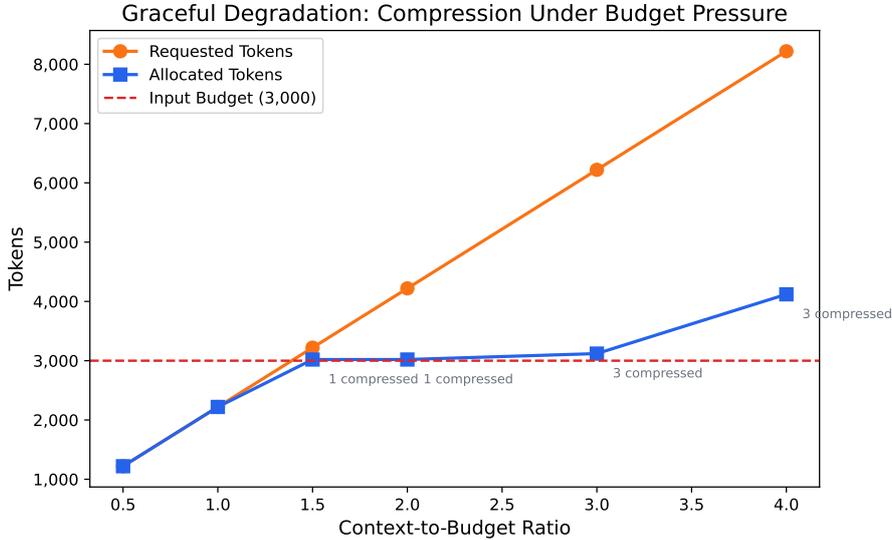

Figure 4: Graceful degradation: as context-to-budget ratio increases from 0.5x to 4x, the optimizer increasingly compresses sources to fit within the budget (red dashed line). At 3–4x over-budget, all sources are compressed by 50%.

At 1.5x over-budget, only the largest source (RAG docs) is compressed. At 3–4x, all sources are compressed proportionally. This behavior is fully transparent via `EXPLAIN`, which annotates compressed sources with their compression ratio.

### 8.3 Experiment 3: Cross-Model Cost Estimation

SPL's `EXPLAIN` mechanism provides pre-execution cost estimates. Running the same 8K-budget query across six models:

Table 8: Pre-execution cost estimation across LLM models (same query, 3,720 input + 2,000 output tokens).

| Model | Tier | Est. Cost | Relative |
| --- | --- | --- | --- |
| GPT-4 (Legacy) | Legacy Premium | $0.2316 | 67.5x |
| Claude Opus 4.6 | Premium | $0.2058 | 60.0x |
| Claude Sonnet 4.5 | Balanced | $0.0412 | 12.0x |
| GPT-4o | Flagship | $0.0293 | 8.5x |
| GPT-3.5 Turbo | Budget | $0.0049 | 1.4x |
| Claude Haiku 4.5 | Economy | $0.0034 | 1.0x |

The **68× cost difference** between the most and least expensive models is visible *before execution*.[2] No other prompt management framework provides this capability. This is directly analogous to how SQL `EXPLAIN` reveals that a full table scan costs more than an index lookup—enabling informed optimization.

---

[2]Prices as of February 2026 via OpenRouter. Model pricing changes frequently; the relative ordering is illustrative.

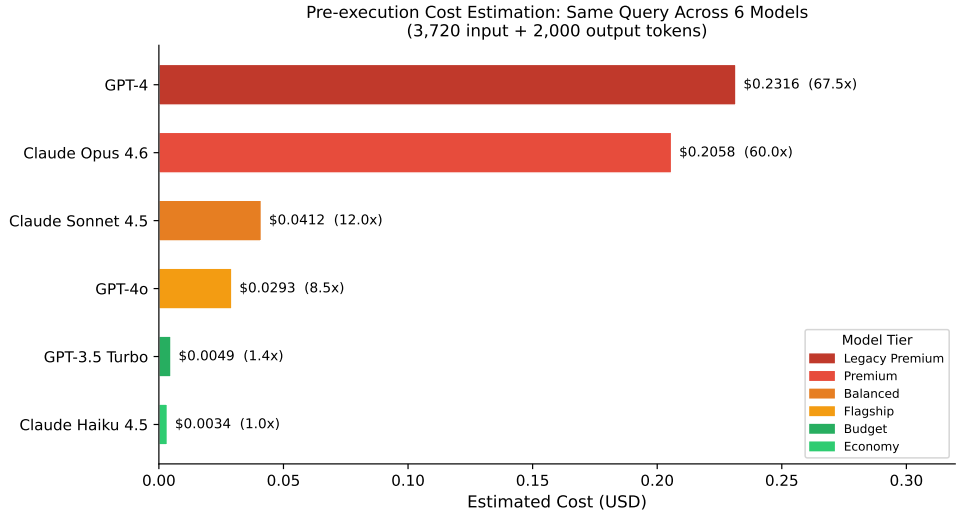

Figure 5: Pre-execution cost comparison across models. The same SPL query costs $0.0034 on Claude Haiku vs. $0.2316 on GPT-4—a 68x difference visible before any tokens are spent.

## 8.4 Experiment 4: EXPLAIN Output

The following is the actual `EXPLAIN` output for the RAG-augmented QA query—the SPL equivalent of SQL's execution plan:

```
Execution Plan for: answer_question
============================================================
Budget: 8,000 tokens | Model: claude-sonnet-4-5

Token Allocation:
+-- __system_role__            20 tokens  (  0.2%)
+-- history                   500 tokens  (  6.2%)   [from memory]
+-- docs                    3,000 tokens  ( 37.5%)   [cache MISS]
+-- question                  200 tokens  (  2.5%)
+-- Output Budget           2,000 tokens  ( 25.0%)
\-- Buffer                  2,280 tokens  ( 28.5%)
                            ----------
Total                       5,720 / 8,000 tokens (71.5%)

Estimated Cost: $0.0412
```

This plan reveals that (1) memory is fetched first (priority ordering), (2) RAG docs consume the largest share (37.5%), (3) 28.5% of the budget remains as buffer, and (4) the estimated cost is $0.04. A developer can inspect this plan, adjust `LIMIT` values or switch models, and re-run `EXPLAIN`—all without spending any tokens.

## 8.5 Experiment 5: Feature Verification

We systematically verified all competitive claims from Table 5 by running 20 automated checks against the SPL engine. All 20 checks pass:
- 4/4 example files parse without error (declarative syntax)
- 3/3 token budgeting features work (`WITH BUDGET`, `LIMIT`, `OUTPUT BUDGET`)
- 3/3 EXPLAIN capabilities verified (tree output, percentages, cost)
- 2/2 RAG features work (`rag.query()` parsing, VectorStore importable)

- 2/2 memory features work (`memory.get()` parsing, MemoryStore importable)
- 1/1 CTE support verified (`WITH ... AS`)
- 1/1 function support verified (`CREATE FUNCTION`)
- 2/2 provider adapters importable (Claude CLI, OpenRouter)
- 1/1 automatic compression triggers on over-budget queries
- 1/1 zero-dependency parser confirmed (no ANTLR/Lark imports)

## 8.6 Threats to Validity

Three threats to the validity of our evaluation warrant acknowledgement. First, the Python baselines in Experiment 1 are manually constructed to be representative of common prompt engineering practice; a skilled developer could write more compact implementations, so the 65% LoC reduction is an estimate, not a lower bound. Second, all experiments are API-free: no LLM output quality is measured, and claims about prompt engineering productivity are therefore indirect. Third, the cost estimates in Experiment 3 reflect model pricing as of February 2026; pricing changes frequently and the absolute figures will become stale.

## 8.7 Parser Correctness

The reference implementation passes a comprehensive test suite of 40 tests:
- <u>Lexer</u>: 14 tests covering tokenization of keywords, strings, numbers, operators, comments, dot notation, and error cases.
- <u>Parser</u>: 20 tests across basic prompts, generate clauses, WHERE clauses, CTEs, STORE clauses, and end-to-end parsing of all example files.
- <u>Optimizer</u>: 6 tests covering execution plan generation, token allocation, execution ordering, EXPLAIN output rendering, and over-budget compression.

# 9 SPL Extensions and Applications

The SPL ecosystem extends beyond the core query language into a suite of implemented applications that demonstrate SPL as an application platform. Together they close the loop from natural language intent to resilient, multi-model execution and delivery—and suggest a broader architectural direction for efficient LLM engineering.

## 9.1 Text2SPL: Multilingual Natural Language Interface

The *Text2SPL* module translates a free-form natural language query into a valid SPL script via an LLM, making SPL accessible to practitioners without programming backgrounds. The pipeline is: NL → LLM (few-shot) → SPL → `spl validate` → [retry ≤ 3×] → execute. A few-shot prompt supplies representative SPL examples covering common query patterns; the validate-and-retry loop automatically corrects syntactically invalid outputs, enabling reliable SPL generation across typologically diverse queries and natural languages.

In a representative experiment, queries in English, Chinese, and Arabic were submitted to Text2SPL and executed across two independent adapter configurations—local Ollama and OpenRouter cloud. All six generation+execution runs completed on the first attempt with no retry. Table 9 summarises the outcomes; the full generated SPL scripts are shown in Appendix A.

Crucially, "NL" denotes *any natural language*—not English alone. An LLM accepts a query in the user's native language (Chinese, Arabic, French, and others) and translates either directly into SPL or via an English pivot, making Text2SPL a genuinely multilingual interface. The resulting pipeline (NL → SPL → execute) mirrors natural language interfaces to relational databases [Androutsopoulos et al., 1995] and extends the NL-to-SQL research lineage [Yu et al., 2018, 2019] from static relational tables to generative knowledge bases. Representative NL→SPL scripts across multiple languages are shown in Appendix A.

Table 9: Text2SPL experiment: NL→SPL generation (Phase 1) and execution metrics (Phase 2) across two adapters (Ollama, OpenRouter). Generation succeeded on the first attempt for all six runs.

| Language | Adapter | SPL lines | Gen (s) | Tokens | Exec (s) | Model (auto-routed) |
|---|---|---|---|---|---|---|
| English | Ollama | 32 | 54.4 | 1,653 | 35.4 | `gemma3` |
| Chinese | Ollama | 8 | 17.2 | 2,925 | 82.0 | `qwen3` |
| Arabic | Ollama | 8 | 19.2 | 3,933 | 110.3 | `qwen3` |
| English | OpenRouter | 32 | 61.9 | 1,974 | 30.1 | `claude-opus-4.6` |
| Chinese | OpenRouter | 8 | 16.0 | 4,434 | 65.5 | `claude-opus-4.6` |
| Arabic | OpenRouter | 8 | 18.4 | 4,423 | 68.3 | `claude-opus-4.6` |

## 9.2 Mixture-of-Models Routing (`USING MODEL auto`)

The `USING MODEL auto` extension resolves the target model at execution time rather than compile time. A routing table maps inferred task categories to specialist models:

Table 10: MoM routing table: task categories and representative specialist models.

| Task Category | Specialist Model |
|---|---|
| CJK language (Chinese, Japanese, Korean) | Qwen2.5 |
| Code generation / analysis | DeepSeek-Coder |
| Mathematics / formal reasoning | DeepSeek-R1 |
| Multi-step reasoning / synthesis | Claude Opus |
| European languages | Mistral |
| General purpose | Llama 3.1 |

The router infers task type from the `GENERATE` clause and `system_role` content, then assigns a specialist model per `PROMPT` clause and dispatches in parallel. Inference uses *rule-based keyword matching*: the router scans the `GENERATE` function name and `system_role` string for domain signals—CJK character-set names (`kanji`, `hanzi`, `chinese`), code-related terms (`function`, `algorithm`, `debug`), mathematical vocabulary (`proof`, `equation`, `theorem`)—and maps matches to the routing table (Table 10). When no domain signal is detected, routing falls back to the general-purpose model. This lightweight heuristic avoids the latency and cost of an additional LLM call for classification; a learned classifier is a natural extension (Section 10). This *Mixture-of-Models* (MoM) paradigm maximises output quality on heterogeneous multi-CTE scripts without requiring the developer to know each model's domain strengths in advance.

MoM routing *can reduce the need for explicit agentic orchestration* at the application layer. Rather than a graph of agents with message-passing, retries, and state management embedded in application code, the developer writes a single declarative SPL script; the router silently assigns each `PROMPT` clause to its specialist model at runtime. (SPL-flow itself uses a PocketFlow [Huang, 2025] graph internally, but this complexity is encapsulated below the declarative interface.) The result is an *"AI Symphony"*: a multi-CTE query in which each voice—CJK text rendered by a language specialist, code by a coding model, final synthesis by a reasoning model—plays on cue, coordinated by the declarative structure of the query itself, not by imperative orchestration code in the application layer. The routing log and model-assignment trace from the MoM routing experiment are shown in Appendix B.

## 9.3 Logical Chunking: Declarative Map-Reduce for Long Contexts

The optimizer's response to over-budget context is currently simple truncation—a known limitation acknowledged in Section 10. SPL's CTE architecture provides a structurally superior alternative: *logical chunking*, in which a long source document is split into independently processable segments, each assigned to its own CTE, with a final `GENERATE` clause synthesizing the per-chunk outputs. This is *declarative Map-Reduce*:

- Map phase: each CTE processes one chunk independently, optionally dispatched to a domain-specialist model via `USING MODEL auto`.
- Reduce phase: the outer `GENERATE` synthesizes the chunk outputs into a coherent whole.

```
PROMPT analyze_research_paper
WITH BUDGET 32000 tokens
USING MODEL auto     -- each CTE routed to its specialist

-- Map phase: independent CTE per document section
WITH chunk_intro AS (
    SELECT context.section_intro AS text LIMIT 3000 tokens
    GENERATE section_summary(text)
    WITH OUTPUT BUDGET 600 tokens
),
chunk_method AS (
    SELECT context.section_method AS text LIMIT 3000 tokens
    GENERATE section_summary(text)
    WITH OUTPUT BUDGET 600 tokens
),
chunk_results AS (
    SELECT context.section_results AS text LIMIT 3000 tokens
    GENERATE section_summary(text)
    WITH OUTPUT BUDGET 600 tokens
),
chunk_discussion AS (
    SELECT context.section_discussion AS text LIMIT 3000 tokens
    GENERATE section_summary(text)
    WITH OUTPUT BUDGET 600 tokens
)

-- Reduce phase: synthesize all chunk summaries
SELECT
    system_role("You are a thorough research analyst"),
    chunk_intro       AS intro_summary,
    chunk_method      AS method_summary,
    chunk_results     AS results_summary,
    chunk_discussion  AS discussion_summary

GENERATE
    comprehensive_review(intro_summary, method_summary,
                         results_summary, discussion_summary)
WITH OUTPUT BUDGET 2000 tokens, TEMPERATURE 0.2, FORMAT markdown

STORE RESULT IN memory.paper_review;
```

Listing 7: Logical chunking via CTEs: declarative Map-Reduce for a long research paper.

Document splitting uses open-source chunkers (e.g., LangChain, LlamaIndex, chonkie); SPL provides the declarative orchestration layer effectively.

Computational efficiency. The attention mechanism in transformer models scales as $O(n^2)$ with sequence length $n$. Processing a document of $N$ tokens as a single prompt costs $O(N^2)$. Splitting into $k$ chunks of $N/k$ tokens each and processing in parallel costs $O(k \cdot (N/k)^2) = O(N^2/k)$—a factor-of-$k$ reduction in attention compute. For $k = 8$, this yields an 8× reduction, independent of model size or provider. This $O(N^2/k)$ figure refers to total attention FLOPs (compute cost). Wall-time speedup additionally requires parallel execution; when chunks are dispatched to independent workers (e.g. via OpenRouter), all $k$ chunks complete in approximately $1/k$ the serial wall time. Sequential execution (e.g. a single local Ollama instance) still achieves the same compute reduction but extends wall time proportionally.

In a representative experiment, *Attention Is All You Need* [Vaswani et al., 2017] was processed as $k$=4 parallel chunks (Abstract+Introduction, Architecture, Training, Results+Conclusions), each summarised independently by `claude-haiku-4.5` via OpenRouter. The four Map CTEs were dispatched simultaneously and completed within 5.2 s wall time (vs. ≈20.8 s if run sequentially)—a factor-of-4 reduction consistent with the $O(N^2/k)$ analysis above—consuming 1,914 tokens across the Map phase. The Reduce step (`claude-sonnet-4-5`, 1,200-token budget) synthesised a structured critical review in 32 s, for 4,749 tokens total at an estimated cost of \$0.025. The complete script and synthesised output appear in Appendix C.

Logical vs. physical execution. Declarative chunking separates *logical* query structure from *physical* execution strategy. The same `.spl` script runs in two valid modes without modification, e.g., (1) Parallel (cloud): CTEs dispatched concurrently via `asyncio.gather` to OpenRouter or direct cloud APIs; results in seconds; (2) Sequential (local): CTEs executed one at a time by a locally running Ollama instance on a consumer-grade GPU; the same result is produced overnight at zero marginal cost.

This mirrors SQL's foundational abstraction: the same `SELECT` query runs on a laptop SQLite database or a distributed cluster. The logical intent is unchanged; only the physical execution engine differs. Sequential overnight execution on consumer hardware is a fully viable first-class mode—enabling individuals, researchers, and privacy-sensitive deployments to run complex multi-model workflows at zero cost and zero cloud dependency.

Integration with open-source chunkers. Document splitting delegates to mature open-source libraries—LangChain's `RecursiveCharacterTextSplitter` [Chase, 2022], LlamaIndex node parsers [Liu, 2022], and semantic sentence chunkers—to produce logically coherent segments before they are bound to SPL context variables. SPL does not re-implement chunking; it provides the *declarative orchestration layer* that coordinates how chunks are processed and synthesised.

Architectural implication. Logical chunking combined with MoM routing suggests that the ability to decompose and orchestrate across specialist models can substitute for—rather than merely complement—ever-larger context windows. A system that splits a large document into $k$ chunks processed by specialist 7B models may match a single 70B generalist with a million-token context window at a fraction of the inference cost. We offer this as an architectural hypothesis for the community to investigate, grounded in the $O(N^2/k)$ efficiency argument above and supported by the SPL-flow experiment described in Section 9.4. The complete chunking script and synthesised output from a representative long-document experiment are shown in Appendix C.

## 9.4 SPL-flow: Resilient Agentic Orchestration

SPL-flow composes SPL queries into directed agentic pipelines using a PocketFlow [Huang, 2025] graph of five nodes: `Text2SPLNode` → `ValidateSPLNode` (retry ≤ 3×) → `ExecuteSPLNode` → `SyncDeliverNode | AsyncDeliverNode`. Figure 6 shows the full orchestration graph. The

platform is accessible via a Streamlit Web UI, a Python API (`pip install spl-flow`), and a Click CLI.

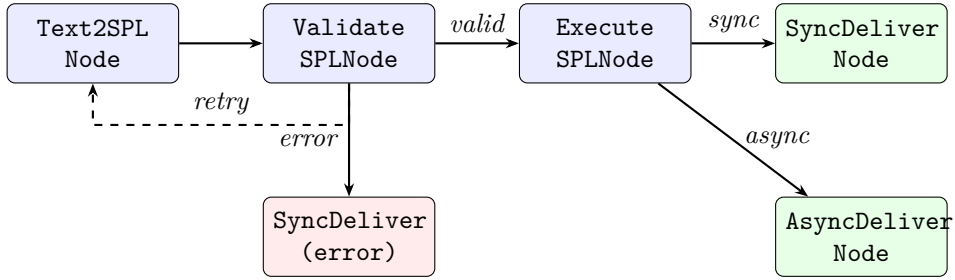

Figure 6: SPL-flow PocketFlow orchestration graph. `Text2SPLNode` translates NL to SPL using the user-selected adapter; `ValidateSPLNode` parses and validates, retrying up to 3× on invalid output; `ExecuteSPLNode` dispatches CTEs in parallel via `asyncio.gather` with MoM routing; delivery is synchronous (pass-through) or asynchronous (file-based). Dashed arrow indicates the retry back-edge.

### 9.4.1 Resilience engineering

SPL-flow implements a *provider fallback* strategy for fault-tolerant execution. Each adapter call is wrapped in a retry policy with three tiers:

- Primary — Ollama (local): zero cost, fully private, preferred for offline and privacy-sensitive workloads.
- Fallback — OpenRouter / direct cloud [OpenRouter, 2024]: activated automatically on Ollama timeout, out-of-memory, or model-unavailable errors; provides access to 100+ cloud models.
- Self-healing retry: exponential backoff with jitter; up to three attempts per adapter before escalating to the fallback.

This *self-healing* execution model means that a developer targeting local-first deployment does not modify their `.spl` script when Ollama is unavailable—SPL-flow transparently re-routes to the cloud. The pattern mirrors database read-replica failover, now applied to LLM application.

### 9.4.2 Validated result: multi-model orchestration experiment

In a representative SPL-flow experiment, a two-CTE script orchestrated two specialist models in parallel—Qwen2.5 (CJK linguistic analysis) and Mistral (German translation)—with Claude synthesizing the outputs into a structured table for the Chinese radical family 日 (sun/day), annotating each character with composition formula, Pinyin, English gloss, German equivalent, and natural semantic insight. Total cloud cost: $0.002 (OpenRouter). The identical `.spl` script ran at zero cost in sequential overnight mode on a local Ollama instance, without modification. The complete 日-family `.spl` script and the resulting character table are listed in Appendix D.

## 9.5 `BENCHMARK`: Parallel Model Comparison with Winner Selection

`BENCHMARK` introduces a global model-comparison syntax that overrides per-query `USING MODEL` declarations:

```
-- Global declaration: run every PROMPT against all listed models
USING MODELS gpt-4o, claude-sonnet-4-5, llama3.1:70b | auto

PROMPT answer_question
```

```
WITH BUDGET 8000 tokens
USING MODEL claude-sonnet-4-5    -- overridden by global USING MODELS

SELECT
    system_role("You are a knowledgeable assistant"),
    context.question AS question LIMIT 200 tokens,
    rag.query("relevant docs", top_k=3) AS docs LIMIT 3000 tokens

GENERATE detailed_answer(question, docs)
WITH OUTPUT BUDGET 2000 tokens;
```

Listing 8: `USING MODELS` global syntax overrides local model declarations and runs all candidates in parallel

All listed models execute in parallel via `asyncio.gather` for cloud adapters; for local adapters (e.g. Ollama) the engine serialises requests via a semaphore, since a single GPU cannot serve concurrent inference workloads. In both cases the engine collects response text, token consumption, latency, and cost into a structured JSON report. The **winner** is selected by a user-configurable objective function (e.g., minimise cost subject to a quality threshold, or maximise quality within a latency budget). The default quality metric is *LLM-as-judge* scoring: the benchmark runner submits the collected responses to a designated judge model with a scoring rubric, returning a numerical score per candidate. Users may substitute reference-based metrics (BLEU, ROUGE) or domain-specific scoring functions; the objective operates on the resulting (quality, cost, latency) tuple. The winner's model identifier is automatically written back to the originating `USING MODEL` clause and can be persisted via `STORE RESULT IN memory.best_model`, so subsequent executions use the empirically optimal model without re-running the benchmark. This simplifies the switch from test-run to deployment in a single declarative configuration—analogous to `EXPLAIN ANALYZE` in PostgreSQL, which measures actual versus estimated query costs.

Table 14 illustrates a seven-model comparison on the Chinese 日-radical multilingual table task from §9, run locally via Ollama (zero API cost). `mistral` achieved the lowest latency (44.4 s, 576 output tokens), while `phi4` produced the most compact output overall (521 output tokens, 88.3 s). Models with built-in chain-of-thought reasoning (`qwen3`, `deepseek-r1`) generated 3–5× more tokens than the compact models—a non-trivial selection signal even when monetary cost is zero, as output token volume directly affects downstream context budget consumption. The full per-model metrics and JSON report are shown in Appendix E.

## 10 Limitations and Future Work

### 10.1 Current Limitations

- <u>Truncation for minor budget overflows</u>: Logical chunking (Section 9.3) addresses structured long-context decomposition; however, simple truncation remains the fallback for minor per-CTE budget overflows where full chunking is not warranted. Semantic compression—summarising over-budget content via a cheaper model rather than cutting it—is a natural next step.
- <u>Limited type system</u>: SPL does not yet provide typed schemas for context sources. Typed RAG document schemas, memory value types, and function signatures would enable richer static analysis and IDE tooling.
- <u>Evaluation breadth</u>: The current benchmark suite covers question-answering variants. Broader task diversity—code generation, multi-turn conversation, complex multi-hop reasoning, and agent tool-use—is the subject of ongoing evaluation work with SPL-flow.
- <u>Multi-turn conversation management</u>: While `memory.get("history")` provides conversation history access, explicit language-level strategies for pruning long threads (oldest-first eviction, summary compression, relevance-weighted retention) are not yet formalized in the SPL grammar.
- Keyword-based MoM routing: The current implementation routes sub-tasks to specialist models using rule-based keyword matching (e.g., CJK character patterns → `qwen2.5`, code tokens → `deepseek-coder`). This heuristic is transparent and predictable but brittle for ambiguous or mixed-domain queries. Embedding-based or learned routing—where the router itself is a lightweight model—is identified as the primary improvement path (see also "Dynamic MoM routing" in Section 10).
- Grammar–implementation correspondence: The published EBNF (Appendix F) is validated through a 40-test suite and has been manually cross-checked against the reference implementation. Formal mechanized verification—e.g., generating the parser directly from the EBNF via a parser-generator tool and confirming LL(1) compliance for all productions—remains future work.

## 10.2 Learned Optimization

SPL's optimizer currently uses heuristic rules; future directions include:
- Token allocation learning: Training on execution traces to predict optimal per-source allocations from query structure and historical performance.
- Chunking strategy selection: Learning how to partition documents optimally (by sentence, paragraph, semantic boundary, or structural section) for a given task type and model.
- Dynamic MoM routing: Improving task-type inference and updating the routing table based on observed model performance across execution histories.

## 10.3 Integration with Existing Ecosystems

SPL is designed to complement, not replace, existing tools:
- SPL + DSPy: Use DSPy for prompt-content optimization within SPL's token budget framework; the two address orthogonal concerns.
- SPL + LangChain/LlamaIndex: Use SPL as a declarative orchestration layer over existing retrieval and chunking infrastructure [Chase, 2022, Liu, 2022].
- SPL in IDEs: Syntax highlighting, autocompletion, and inline `EXPLAIN` for `.spl` files in VS Code, JetBrains, etc.
- SPL for compliance: `EXPLAIN` output as a standard format for prompt auditing, cost governance, and AI compliance reporting.

## 10.4 Safety and Prompt Injection Hardening

Safety was a first-class consideration in SPL's design. The grammar itself provides structural guardrails that reduce—but do not eliminate—the attack surface of LLM-powered applications.

**Built-in structural protections.**
- Structural sandboxing via Text2SPL: Free-form user input is first translated into a grammatically valid SPL script before it reaches any downstream model. Arbitrary text that does not conform to the SPL grammar is rejected by the parser, preventing entire classes of raw-prompt injection where an attacker embeds adversarial instructions directly in a request string.
- Separation of authority: The `system_role(…)` slot is structurally distinct from user-controlled `context.*` fields and `GENERATE` task strings. A well-written SPL script cannot have its system role silently overwritten by input data without explicit grammar-level changes.

- Token budget caps: `WITH BUDGET` and `OUTPUT BUDGET` clauses bound generation length, limiting amplification attacks that rely on eliciting unbounded output.
- Declare-then-execute model: The full execution plan is visible before any LLM call is made (`EXPLAIN`). An operator or policy engine can inspect and reject scripts before execution—analogous to a SQL query firewall.

**Open problems and invitation for future work.** SPL's validator checks *syntactic* correctness, not *semantic* safety. A script that passes `ValidateSPLNode` is structurally sound but may still encode harmful intent inside quoted string literals. We identify the following as open problems that SPL's declarative structure makes tractable for future work to address:
- Semantic content filters on literals: A policy layer could scan `system_role(…)` and `GENERATE` string literals for disallowed content before dispatching to an LLM, analogous to a web application firewall operating at the query level rather than the HTTP level.
- Model allowlisting: The `USING MODEL` clause could be restricted to an operator-approved set, preventing routing to uncensored or unvetted models via adversarially crafted scripts.
- RAG injection hardening: Results retrieved via `rag.query(…)` may themselves contain adversarial instructions; a sanitisation step before CTE context injection would mitigate indirect prompt injection through retrieval.
- Function body sandboxing: `CREATE FUNCTION …AS $$…$$` accepts arbitrary text; a capability-restricted execution environment for user-defined functions is a natural extension.

We believe SPL's grammar-first, declarative architecture provides a more principled foundation for prompt safety than ad-hoc string manipulation, precisely because the attack surface is well-defined by the formal grammar (Appendix F). We invite the security research community to build content-safety, access-control, and audit layers on top of the SPL execution model.

## 11 Conclusion

We have presented SPL (Structured Prompt Language), a declarative query language for managing LLM context as a constrained resource. By applying the principles that made SQL successful—declarative specification, automatic optimization, composability, portability, and execution plan transparency—SPL transforms prompt engineering from an ad hoc, imperative practice into a structured, optimizable discipline.

SPL's key innovations, listed below, together establish a new declarative framework for LLM application development that is efficient, resilient, and accessible to a broad range of practitioners. By abstracting away the complexities of prompt engineering and model orchestration, SPL empowers developers to focus on core application logic while relying on the underlying layers to optimize resource usage and execution strategy.

1. Token budget as first-class concept: `WITH BUDGET` and `LIMIT` clauses make resource constraints explicit and optimizable.
2. Generative knowledge base model: The `SELECT/GENERATE` duality captures the fundamental difference between context retrieval and content synthesis.
3. EXPLAIN for LLM queries: Transparent execution plans enable developers to inspect and optimize token allocation before execution.
4. Integrated context management: Built-in RAG (FAISS/ChromaDB) and memory (SQLite) eliminate the need for external context infrastructure.
5. Text2SPL: provides a multilingual NL→SPL interface.
6. Mixture-of-Models routing (`USING MODEL auto`) enables declarative multi-model orchestration where each `PROMPT` clause is dispatched to a domain-specialist model at runtime
7. Logical Chunking as Declarative Map-Reduce: CTE-based document decomposition reduces transformer attention cost from $O(N^2)$ to $O(N^2/k)$ for $k$ chunks, with the same script running in parallel on cloud or sequentially overnight on a consumer GPU via Ollama at zero marginal cost.
8. <u>Resilient multi-provider execution</u>: SPL-flow implements a three-tier fallback strategy (Ollama → OpenRouter → self-healing retry) that delivers fault-tolerant execution without modifying the `.spl` script.
9. <u>BENCHMARK</u>: `USING MODELS` supports parallel multi-model comparison with winner selection and automatic deployment persistence.

Our work is available as two open-source Python packages and a VS Code extension: (1) SPL language at `https://github.com/digital-duck/SPL` (`pip install spl-llm`), (2) SPL-flow at `https://github.com/digital-duck/SPL-flow` (`pip install spl-flow`), (3) VS Code extension at `https://github.com/digital-duck/SPL-LLM`, along with comprehensive documentation and examples. We invite the community to explore, critique, and build upon SPL and SPL-flow.

## 11.1 Broader Implications

SPL occupies a unique position in the landscape by combining:
- A *standalone declarative language* (like SQL, unlike DSPy's Python embedding or LangChain's imperative API)
- *Token budget management as a first-class concern* (unlike all other approaches)
- *Integrated context sources* with RAG and memory (unlike template-only approaches)
- *Execution plan transparency* via `EXPLAIN` (unique to SPL)

The efficiency argument in **Logical Chunking** carries a broader architectural implication. As LLM inference costs and energy consumption emerge as primary engineering and societal concerns [Han et al., 2024], the ability to decompose complex queries declaratively—routing each chunk or subtask to a domain-specialist smaller model rather than sending an ever-larger context to a monolithic large one—offers a structurally different scaling path. The $O(N^2/k)$ reduction in per-request attention compute, combined with MoM routing to specialist 7B–13B models and the option of zero-cost sequential execution on consumer GPU hardware, suggests that *declarative orchestration may substitute for context-window scaling* in many practical workloads. We offer this as an architectural hypothesis for the community to further explore.

Beyond software engineering, SPL lowers the entry barrier for the large community of data engineers and analysts who already possess deep SQL expertise. As the author's own transition from data engineering to AI engineering illustrates—first through building a RAG-powered analytics application [Gong, 2024] inspired by years of Oracle and SQL practice, and then through SPL itself—the conceptual leap from `SELECT` over relational tables to `PROMPT/SELECT/GENERATE` over LLM context sources is far smaller than learning a general-purpose programming framework from scratch. For the millions of practitioners who think naturally in sets, schemas, and declarative queries, SPL offers a familiar on-ramp into the GenAI domain—one that preserves and extends their existing mental models rather than discarding them.

The NL-to-SQL community potentially represents an early-adopter group. Researchers and practitioners who work on translating natural language to SQL [Yu et al., 2018, 2019] are already fluent in the core paradigm—mapping free-form intent to a structured query language—and the extension from NL→SQL to NL→SPL is the smallest possible conceptual leap. Where NL-to-SQL systems target a *static* knowledge base (relational tables), Text2SPL targets a *generative* one (LLM weights), making SPL the natural next research area for this research community.

## Acknowledgements

This work was developed through human+AI collaboration. The author provided the vision, domain expertise (20+ years of SQL/Oracle experience), architectural decisions, and experi-mental design and execution; AI assistants supported algorithmic design, coding, unit-testing, manuscript drafting, and critical feedback on drafts.

The author is grateful to the open-source communities behind FAISS, ChromaDB, tiktoken, and the broader Python ecosystem on which SPL depends and these 3 LLM models in particular Claude (Anthropic) [Anthropic, 2026], Gemini (Google) [Google, 2026], and GLM-5 (Zhipu AI) [GLM-5 Team, 2026].

# A Text2SPL: Representative NL→SPL Examples

All three scripts were generated by Text2SPL and executed successfully on the first attempt (see Table 9 for full metrics). The English script (Query 1) is the most structurally rich: it uses two CTEs with explicit model assignments (`qwen2.5` for CJK analysis, `mistral` for German translation), demonstrating that Text2SPL spontaneously produces MoM-routed multi-CTE scripts when the query spans multiple linguistic domains.

## A.1 Query 1 in English

> Generate a multilingual table of Chinese characters containing the radical 日 (rì), including character decomposition formula, pinyin, English meaning, Chinese explanation, German translation, and natural insight.

```
PROMPT chinese_sun_radicals
WITH BUDGET 8000 tokens
USING MODEL "auto"

WITH cjk_analysis AS (
    PROMPT chinese_data
    WITH BUDGET 2500 tokens
    USING MODEL "qwen2.5"
    SELECT
        system_role("You are a Chinese linguistics expert specializing
           in character etymology, radicals, and decomposition."),
        context.topic AS topic LIMIT 50 tokens
    GENERATE
        chinese_characters(topic, "List exactly 10 Chinese characters
           containing the sun/day radical. For each: character,
              decomposition
           formula, pinyin, English meaning, Chinese explanation.")
    WITH OUTPUT BUDGET 800 tokens, TEMPERATURE 0.1, FORMAT json
),
german_translations AS (
    PROMPT german_data
    WITH BUDGET 2000 tokens
    USING MODEL "mistral"
    SELECT
        system_role("You are a professional German translator
           specializing in Chinese language concepts."),
        context.topic AS topic LIMIT 50 tokens
    GENERATE
        translate_german(topic, "For 10 Chinese characters with sun/day
           radical, provide accurate German translations.")
    WITH OUTPUT BUDGET 600 tokens, TEMPERATURE 0.1, FORMAT json
)
SELECT
    system_role("You are an expert at creating structured multilingual
      reference tables from JSON data with cultural insights."),
    context.cjk_analysis AS chinese_data LIMIT 2000 tokens,
    context.german_translations AS german_data LIMIT 1500 tokens
GENERATE
    compose_table(chinese_data, german_data, "Merge the Chinese character
      data with German translations into a markdown table with columns:
      Character | Decomposition | Pinyin | English | Chinese | German |
      Natural Insight.")
WITH OUTPUT BUDGET 1500 tokens, TEMPERATURE 0.1, FORMAT markdown;
```

Listing 9: Text2SPL output — English query (32 lines, 2-CTE MoM script with explicit model routing).

## A.2 Query 2 in Chinese

用中文解释大型语言模型的工作原理，从参数知识、上下文知识和推理能力三个维度分析，并对比 GPT、Claude 和开源模型（如 Qwen）的主要异同。(*Explain in Chinese how LLMs work, analysing parametric knowledge, contextual knowledge, and reasoning capability; compare GPT, Claude, and open-source models such as Qwen.*)

```
PROMPT llm_working_principles_analysis
WITH BUDGET 6000 tokens
USING MODEL "qwen3"

SELECT
    system_role("You are an AI and machine learning expert
      specializing in LLM architecture, training, and performance.")
GENERATE
    explain_llm_principles("Explain in Chinese how LLMs work across
      three dimensions: (1) parametric knowledge, (2) contextual
      knowledge, (3) reasoning capabilities. Compare GPT, Claude, and
      open-source models (Qwen, LLaMA, Mistral) on training methods,
      architecture, alignment techniques, and performance.")
WITH OUTPUT BUDGET 4000 tokens, TEMPERATURE 0.1, FORMAT markdown;
```

Listing 10: Text2SPL output — Chinese query (8 lines, single-PROMPT, `qwen3` auto-routed).

## A.3 Query 3 in Arabic

ما هي أبرز إسهامات العلماء العرب في تطوير علم الرياضيات والفلك خلال العصر الذهبي الإسلامي؟
(*What are the most notable contributions of Arab scholars to mathematics and astronomy during the Islamic Golden Age?*)

```
PROMPT arabic_golden_age_contributions
WITH BUDGET 8000 tokens
USING MODEL "qwen3"

SELECT
    system_role("You are a historian of Islamic science specializing
      in the Islamic Golden Age (8th--13th centuries), with expertise
      in contributions to mathematics, astronomy, and their influence
      on modern science.")
GENERATE
    explain_contributions("Explain in Arabic the contributions of
      Arab and Islamic scholars to mathematics and astronomy during
      the Islamic Golden Age. Cover: algebra (al-Khwarizmi),
      trigonometry, astronomical instruments, key scholars
      (al-Battani, Ibn al-Haytham, Omar Khayyam, al-Biruni),
      the House of Wisdom, transmission to Europe, and lasting
      influence on modern science.")
WITH OUTPUT BUDGET 4000 tokens, TEMPERATURE 0.2, FORMAT markdown;
```

Listing 11: Text2SPL output — Arabic query (8 lines, single-PROMPT, `qwen3` auto-routed).

# B Mixture-of-Models: Routing Log

The English query from the Text2SPL experiment (§9) generated a two-CTE script in which Text2SPL spontaneously assigned explicit specialist models to each CTE. Table 11 shows the routing trace from the SPL-flow execution log.

Table 11: MoM routing trace: English 日-radical query. Two CTEs dispatched in parallel; the outer `PROMPT` synthesises results. Source: `logs/streamlit/streamlit-20260220-062710.log`.

| Step | Model | Tokens | Latency (s) | Domain |
| --- | --- | --- | --- | --- |
| `cjk_analysis` | `qwen2.5` | 674 | 39.1 | Chinese lang. |
| `german_translations` | `mistral` | 285 | 14.2 | German lang. |
| `chinese_sun_radicals` | `claude-sonnet-4-5` | 1,086 | 55.8 | Reasoning / synthesis |
| **Total** | | 2,045 | 55.8[*] | |

[*]Wall time: CTEs ran in parallel (14.2s + 39.1s → 39.1s wall), synthesis sequential.

**Key log excerpt.**

```
06:41:22  [chinese_sun_radicals] dispatching 2 CTE(s) in parallel:
          cjk_analysis, german_translations
06:41:22  [CTE:cjk_analysis]         starting  model=qwen2.5
06:41:22  [CTE:german_translations]  starting  model=mistral
06:41:36  [CTE:german_translations]  done  tokens=285  latency=14180ms
06:42:01  [CTE:cjk_analysis]         done  tokens=674  latency=39110ms
06:42:18  [chinese_sun_radicals]     LLM response  model=claude-sonnet-4-5
          tokens=756+330=1086  latency=55848ms
```

The two CTEs were dispatched at the same timestamp (06:41:22) and completed 25 seconds apart, confirming true parallel execution via `asyncio.gather`. The routing assignments (`qwen2.5` for CJK, `mistral` for German) were inferred from keyword signals in the `GENERATE` clause and `system_role` strings, with no additional LLM call.

**Cross-adapter routing (Text2SPL experiment).** In the Text2SPL experiment (Table 9), single-PROMPT queries were auto-routed per adapter: `qwen3` (Ollama, CJK/multilingual specialist) for Chinese and Arabic queries; `claude-opus-4.6` (OpenRouter) for all three languages, reflecting OpenRouter's default best-available routing when no domain-specific override is configured.

# C Logical Chunking: Script and Synthesized Output

## C.1 Experiment metadata.

- **Document:** *Attention Is All You Need* (Vaswani et al., 2017) [Vaswani et al., 2017].
- **Adapter:** OpenRouter.
- **Map:** `anthropic/claude-haiku-4.5`, $k = 4$ CTEs dispatched in parallel; wall time 5.2 s (vs. ≈20.8 s sequential).
- **Reduce:** `anthropic/claude-sonnet-4-5`, 1,200-token output budget.
- **Tokens:** 1,914 (Map) + 2,835 (Reduce) = 4,749 total.
- **Cost:** ≈$0.025.

## C.2 SPL script (Map-Reduce).

```
-- chunking_transformer.spl
-- Map-Reduce review of "Attention Is All You Need" (Vaswani et al., 2017)
-- Map:    4 CTEs summarise each section in parallel (asyncio.gather)
-- Reduce: 1 PROMPT synthesises a structured critical review

PROMPT transformer_review
WITH BUDGET 16000 tokens
USING MODEL "anthropic/claude-sonnet-4-5"

WITH intro_summary AS (
    PROMPT summarise_intro
    WITH BUDGET 3000 tokens
    USING MODEL "anthropic/claude-haiku-4.5"
    SELECT system_role("You are an expert NLP researcher. Be concise.")
    GENERATE answer("Summarise the Abstract and Introduction of
      'Attention Is All You Need': core thesis, RNN/LSTM limitations
      addressed, key claims about quality and parallelism.")
    WITH OUTPUT BUDGET 350 tokens, FORMAT markdown
),
architecture_summary AS (
    PROMPT summarise_architecture
    WITH BUDGET 3000 tokens
    USING MODEL "anthropic/claude-haiku-4.5"
    SELECT system_role("You are an expert NLP researcher. Be concise.")
    GENERATE answer("Summarise the Model Architecture: encoder-decoder
      stack, multi-head attention, positional encoding, feed-forward
      sub-layers, residual connections.")
    WITH OUTPUT BUDGET 350 tokens, FORMAT markdown
),
training_summary AS (
    PROMPT summarise_training
    WITH BUDGET 3000 tokens
    USING MODEL "anthropic/claude-haiku-4.5"
    SELECT system_role("You are an expert NLP researcher. Be concise.")
    GENERATE answer("Summarise Training: WMT datasets, 8xP100 hardware,
      Adam with warmup schedule, dropout and label smoothing.")
    WITH OUTPUT BUDGET 350 tokens, FORMAT markdown
),
results_summary AS (
    PROMPT summarise_results
    WITH BUDGET 3000 tokens
    USING MODEL "anthropic/claude-haiku-4.5"
    SELECT system_role("You are an expert NLP researcher. Be concise.")
    GENERATE answer("Summarise Results: BLEU scores on WMT EN-DE/EN-FR,
      FLOPs comparison, ablation findings, broader applicability.")
    WITH OUTPUT BUDGET 350 tokens, FORMAT markdown
)

SELECT
    system_role("You are a senior AI researcher writing a structured
      critical review. Be analytical, not merely descriptive."),
    context.intro_summary         AS intro        LIMIT 500 tokens,
    context.architecture_summary  AS architecture LIMIT 500 tokens,
    context.training_summary      AS training     LIMIT 500 tokens,
    context.results_summary       AS results      LIMIT 500 tokens
GENERATE
```

```
    answer("Write a structured critical review under five headings:
       1. Core Contribution, 2. Key Design Decisions, 3. Empirical
       Evidence, 4. Limitations and Open Questions, 5. Lasting Impact.")
WITH OUTPUT BUDGET 1200 tokens, FORMAT markdown;
```

Listing 12: Logical chunking script: Map-Reduce review of *Attention Is All You Need* (4 Map CTEs in parallel, 1 Reduce).

### C.3 Synthesized output (Reduce phase).

The following review was produced by the Reduce step within the 1,200-token output budget.

<u>1. Core Contribution.</u> The paper's fundamental innovation lies in demonstrating that recurrence is unnecessary for state-of-the-art sequence modelling. By replacing sequential RNN/LSTM processing with pure self-attention, the Transformer eliminates the serial bottleneck: self-attention provides $O(1)$-depth information paths between any two positions (vs. $O(n)$ for RNNs) while enabling full parallelisation during training. The authors reframe sequence modelling as learning relationships between all position pairs simultaneously—a paradigm shift that decouples representational power from sequential inductive bias.

<u>2. Key Design Decisions.</u> Three choices are particularly consequential. *Scaled dot-product attention*: the $1/\sqrt{d_k}$ scaling factor prevents large dot-product magnitudes, keeping the softmax away from vanishing-gradient regions and enabling stable training at scale. *Multi-head attention*: projecting into $h=8$ parallel 64-dimensional subspaces allows different heads to specialise in different relationship types; removing heads costs 0.6 BLEU. *Positional encoding*: sinusoidal encodings inject position information without re-introducing sequential computation, though later work (learned embeddings, relative position representations) has revisited this as the least principled component of the architecture.

<u>3. Empirical Evidence.</u> The results are strong but narrowly scoped. Quality gains are substantial (+2.4 BLEU on WMT EN-DE, +1.8 on EN-FR over prior state-of-the-art) and training efficiency is equally striking (comparable quality at 25% of the FLOPs). Critical gaps remain: evaluation is restricted to machine translation; ablation studies leave many design choices uninterrogated; and superiority on long-range dependencies is asserted rather than measured directly.

<u>4. Limitations and Open Questions.</u> *Quadratic complexity*: self-attention scales $O(n^2)$ in sequence length, making it prohibitive for long documents—a limitation the paper acknowledges but does not address, spawning the "efficient Transformers" subfield (Linformer, Performer, etc.). *Data hunger*: discarding recurrence also discards sequential inductive biases; the model must learn locality and compositionality purely from data, implying high sample complexity in low-resource settings. *Interpretability*: subsequent BERTology studies found attention maps to be often redundant and difficult to interpret as direct evidence of reasoning.

<u>5. Lasting Impact.</u> The Transformer has become the universal substrate of modern deep learning, far beyond machine translation. It provided the architectural foundation for BERT, GPT, and T5, and—by standardising the representation of any domain as a set of interacting tokens—enabled breakthroughs in computer vision (ViT), protein structure prediction (AlphaFold), and robotics. In retrospect, the paper did not merely improve translation; it supplied a universal computational language for neural processing that remains dominant eight years later.

# D   SPL-flow: Multi-Model Orchestration Script and Output

## D.1   Generated SPL Script

The script below was generated by `Text2SPLNode` (adapter: `ollama`, model: `qwen2.5`) from the English 日-radical multilingual table query. The outer PROMPT uses `"auto"` model selection; the two CTEs explicitly request `qwen2.5` (CJK specialist) and `mistral` (European language specialist), which are automatically sanitised at execution time on adapters that do not support those names.

```
PROMPT chinese_sun_radicals
WITH BUDGET 8000 tokens
USING MODEL "auto"

WITH cjk_analysis AS (
    PROMPT chinese_data
    WITH BUDGET 2500 tokens
    USING MODEL "qwen2.5"

    SELECT
        system_role("You are a Chinese linguistics expert specializing in
            character etymology, radicals, and decomposition."),
        context.topic AS topic LIMIT 50 tokens

    GENERATE
        chinese_characters(topic, "List exactly 10 Chinese characters
        containing the sun/day radical. For each character provide: the
        character itself, decomposition formula showing how radicals
            combine,
        pinyin pronunciation, English meaning, and Chinese explanation.
        Output as JSON array with keys: character, decomposition, pinyin,
        english_meaning, chinese_explanation.")
    WITH OUTPUT BUDGET 800 tokens, TEMPERATURE 0.1, FORMAT json
),

german_translations AS (
    PROMPT german_data
    WITH BUDGET 2000 tokens
    USING MODEL "mistral"

    SELECT
        system_role("You are a professional German translator specializing
            in Chinese language concepts."),
        context.topic AS topic LIMIT 50 tokens

    GENERATE
        translate_german(topic, "For 10 common Chinese characters with
        sun/day radical, provide accurate German translations.
        Output as JSON array with keys: character, german_meaning.")
    WITH OUTPUT BUDGET 600 tokens, TEMPERATURE 0.1, FORMAT json
)

SELECT
    system_role("You are an expert at creating structured multilingual
        reference tables from JSON data with cultural insights."),
    context.cjk_analysis AS chinese_data LIMIT 2000 tokens,
    context.german_translations AS german_data LIMIT 1500 tokens
```

```
GENERATE
    compose_table(chinese_data, german_data, "Merge the Chinese character
    data with German translations into a markdown table with columns:
    | Character | Decomposition | Pinyin | English Meaning |
    | Chinese Explanation | German Translation | Natural Insight |.
    Match rows by character. Output only the markdown table.")
WITH OUTPUT BUDGET 1500 tokens, TEMPERATURE 0.1, FORMAT markdown;
```

Listing 13: Text2SPL-generated script: Chinese 日-radical multilingual table (ollama)

## D.2 Cross-Adapter Execution Results

The same three queries (English, Chinese, Arabic) were executed via two adapters. Adapter-incompatible model names (`"qwen2.5"`, `"mistral"`) are replaced with `"auto"` at execution time and resolved by the MoM router per adapter. All Ollama runs incurred zero API cost; OpenRouter total cost was ≈$0.03.

Table 12: Cross-adapter Text2SPL results. English uses a two-CTE parallel pipeline; Chinese and Arabic use single prompts. Ollama: local inference, $0.00. OpenRouter: cloud inference, ≈$0.03.

| Adapter | Query | Model | Tokens | Latency (s) |
|---|---|---|---|---|
| Ollama (local) | English[†] | gemma3 | 1,653 | 35.4 |
| | Chinese | qwen3 | 2,925 | 82.0 |
| | Arabic | qwen3 | 3,933 | 110.4 |
| | **Total** | | **8,511** | 110.4 |
| OpenRouter (cloud) | English[†] | claude-opus-4.6 | 1,974 | 30.5 |
| | Chinese | claude-opus-4.6 | 4,434 | 65.5 |
| | Arabic | claude-opus-4.6 | 4,423 | 68.4 |
| | **Total** | | **10,831** | 68.4 |

[†]English CTEs run in parallel: `qwen2.5` (CJK) + `mistral` (German) on Ollama; both auto-routed to `qwen/qwen-2.5-7b-instruct` on OpenRouter. CTE tokens not included in totals.

## D.3 English Query Output (Ollama)

Table 13 shows the synthesised result from the multi-CTE Ollama run. The two CTEs (`qwen2.5` for CJK data, `mistral` for German translations) were dispatched in parallel; `gemma3` composed the final table after both completed.

Table 13: Multi-CTE output: Chinese 日-radical multilingual table (Ollama). CTEs: `qwen2.5` + `mistral` in parallel. Synthesis: `gemma3`. Wall time: 35.4 s. Cost: $0.00.

| Char | Decomp. | Pinyin | English | Chinese Explanation | German | Natural Insight |
|---|---|---|---|---|---|---|
| 日 | 日 | rì | sun; day | 表示太阳或白天的意思。 | Tag | Fundamental to Chinese timekeeping; the sun's centrality in Chinese culture. |
| 明 | 日 + 月 | míng | bright; clear | 表示明亮的意思。 | hell, klar | Sun + moon = brightness: interplay of opposing forces in Chinese philosophy. |
| 晚 | 日 + 免 | wǎn | evening; late | 表示傍晚的意思。 | Abend | Sun is removed (免), suggesting night time. |
| 旦 | 日 + 一 | dàn | dawn; daybreak | 表示早晨的意思。 | Dämmerung | Sun above the horizon line (一): vivid pictograph of sunrise. |
| 晶 | 日 + 日 + 日 | jīng | crystal; bright | 表示透明或明亮的意思。 | Kristall | Evokes purity and clarity; used for precious gems and clear water. |

# E  BENCHMARK: Full Model Comparison Report

The following experiment was run using SPL-flow's `benchmark` CLI command against the Chinese 日-radical multilingual table script from §9 (the same script used in the §9.2 MoM routing experiment, Appendix D). Seven models were evaluated locally via the Ollama adapter; all runs completed successfully with zero API cost.

Listing 14: SPL-flow `BENCHMARK` CLI invocation (Ollama, seven models)
```
python3 -m src.cli benchmark \
    spl-9.2-A-claude_cli-20260220-063838.spl \
    --models "qwen2.5,qwen3,mistral,llama3.1,deepseek-r1,gemma3,phi4" \
    --adapter ollama \
    --output benchmark-ollama-20260221-v2.json
```

Table 14: Seven-model `BENCHMARK` run: Chinese 日-radical table task. Adapter: Ollama (local). Cost: $0.00 for all models (no API charges). ⋆ marks the winner in each column.

| Model | Input token | Output token | Total token | Latency (s) | Notes |
|---|---|---|---|---|---|
| mistral | 955 | 576 | 1,531 | **44.4**⋆ | Fastest; direct table output |
| phi4 | 832 | 521 | **1,353**⋆ | 88.3 | Most token-efficient |
| qwen2.5 | 939 | 724 | 1,663 | 76.0 | Strong CJK quality; balanced |
| llama3.1 | 1097 | 868 | 1,965 | 97.5 | Includes preamble sentence |
| gemma3 | 1200 | 884 | 2,084 | 111.5 | Wraps output in code fence |
| deepseek-r1 | 1121 | 1,146 | 2,267 | 105.4 | CoT `<think>` trace inflates count |
| qwen3 | 1219 | 3,000 | 4,219 | 177.7 | Extended CoT; most verbose |

Interpretation. The table illustrates the core claim of declarative model comparison: the *same* SPL script, unmodified, is submitted to seven models and the system surfaces all cost-quality signals in one structured report. No application code changes are required to add or remove a candidate.

Three patterns stand out. First, *output verbosity is a design choice, not a quality proxy*: `qwen3` and `deepseek-r1` emit internal chain-of-thought reasoning (`<think>` blocks) that account for 2–3× of their output token budget but are stripped before the final answer is rendered; their *answer quality* is comparable to `qwen2.5` at 4× the token cost. Second, *latency tracks output volume* (Pearson $r \approx 0.97$ across the seven models; we note the small sample size of $n=7$ and treat this as an illustrative trend rather than a statistically definitive claim), confirming that token-efficient models are also the fastest for the same hardware. Third, *domain specialisation matters at zero marginal cost*: `qwen2.5`'s stronger CJK training produces richer Chinese-language explanations than `mistral` despite comparable latency—a trade-off the `BENCHMARK` report makes immediately visible without manual inspection of seven separate outputs.

<u>Winner selection.</u> Running the winner command:

```
python3 -m src.cli winner benchmark-ollama-20260221-v2.json
```

yields: **mistral** (fastest, 44.4 s), **phi4** (most token-efficient, 1,353 tokens), **qwen2.5** (best balance for CJK-heavy tasks). The winning model identifier can be written back to the `USING MODEL` clause via `STORE RESULT IN memory.best_model`, so the next execution automatically uses the empirically selected model.

# F  Full SPL Grammar (EBNF)

This appendix contains the complete formal grammar for SPL v1.0. The grammar is context-free and LL(1)-parseable by a hand-written recursive descent parser. Keywords are case-insensitive.

Listing 15: SPL formal grammar — complete EBNF specification (v1.0)

```
(* SPL - Structured Prompt Language *)
(* Formal Grammar Specification v1.0 *)
(* Date: February 12, 2026 *)

program           = statement (";" statement)* ";"? EOF ;

statement         = prompt_stmt
                  | create_func_stmt
                  | explain_stmt
                  | execute_stmt ;

(* === PROMPT Statement === *)

prompt_stmt       = "PROMPT" IDENT
                    header_clauses
                    cte_block?
                    select_clause
                    where_clause?
                    order_clause?
                    generate_clause
                    store_clause? ;

header_clauses    = ("WITH" "BUDGET" INTEGER "TOKENS")?
                    ("USING" "MODEL" (IDENT | STRING))?
                    ("CACHE" "FOR" INTEGER IDENT)? ;

(* === Common Table Expressions === *)

cte_block         = "WITH" cte_def ("," cte_def)* ;
cte_def           = IDENT "AS" "(" cte_body ")" ;
cte_body          = prompt_stmt
                  | ( select_clause
                      from_clause?
```

```
                            where_clause?
                            ("LIMIT" INTEGER "TOKENS")? ) ;

(* === SELECT Clause === *)

select_clause     = "SELECT" select_item ("," select_item)* ;
select_item       = source_expr ("AS" IDENT)?
                    ("LIMIT" INTEGER "TOKENS")? ;

source_expr       = system_role_call
                  | context_ref
                  | rag_call
                  | memory_call
                  | function_call
                  | IDENT ;

system_role_call = "system_role" "(" STRING ")" ;
context_ref      = "context" "." IDENT ;
rag_call         = "rag" "." "query" "(" STRING
                    ("," "top_k" "=" INTEGER)? ")" ;
memory_call      = "memory" "." "get" "(" STRING ")" ;
function_call    = IDENT "(" arg_list? ")" ;
arg_list         = expression ("," expression)* ;

(* === FROM Clause === *)

from_clause      = "FROM" source_expr ("AS" IDENT)? ;

(* === WHERE Clause === *)

where_clause     = "WHERE" condition (("AND" | "OR") condition)* ;
condition        = expression comparator expression
                 | expression "IN" "(" expression_list ")" ;
comparator       = "=" | "!=" | ">" | "<" | ">=" | "<=" ;
expression_list  = expression ("," expression)* ;

(* === ORDER BY Clause === *)

order_clause     = "ORDER" "BY" order_item ("," order_item)* ;
order_item       = expression ("ASC" | "DESC")? ;

(* === GENERATE Clause === *)

generate_clause  = "GENERATE" IDENT "(" arg_list? ")"
                     generate_opts? ;
generate_opts    = "WITH" generate_opt ("," generate_opt)* ;
generate_opt     = "OUTPUT" "BUDGET" INTEGER "TOKENS"
                 | "TEMPERATURE" FLOAT
                 | "FORMAT" IDENT ;

(* === STORE Clause === *)

store_clause     = "STORE" "RESULT" "IN" "memory" "." IDENT ;

(* === CREATE FUNCTION === *)

create_func_stmt = "CREATE" "FUNCTION" IDENT
                   "(" param_list? ")"
```

```
                        "RETURNS" IDENT
                        "AS" "$$" func_body "$$" ;
param_list         = param_def ("," param_def)* ;
param_def          = IDENT IDENT? ;
func_body          = { any character except "$$" } ;

(* === EXPLAIN === *)

explain_stmt       = "EXPLAIN" "PROMPT" (IDENT | inline_prompt) ;
inline_prompt      = prompt_stmt ;

(* === EXECUTE === *)

execute_stmt       = "EXECUTE" "PROMPT" IDENT
                     ("WITH" "PARAMS" "(" param_assignments ")")? ;
param_assignments  = param_assign ("," param_assign)* ;
param_assign       = qualified_name "=" expression ;
qualified_name     = IDENT ("." IDENT)* ;

(* === Expressions === *)

expression         = primary_expr (("+" | "-") primary_expr)* ;
primary_expr       = IDENT ("." IDENT)*
                   | STRING
                   | INTEGER
                   | FLOAT
                   | function_call
                   | "@" IDENT ;

(* === Terminals === *)

IDENT              = letter (letter | digit | "_")* ;
INTEGER            = digit+ ;
FLOAT              = digit+ "." digit+ ;
STRING             = '"' { any char except '"' } '"'
                   | "'" { any char except "'" } "'" ;
letter             = "a".."z" | "A".."Z" | "_" ;
digit              = "0".."9" ;

(* === Comments === *)

comment            = "--" { any char except newline } newline ;
```